\documentclass{article} 
\usepackage{iclr2026_conference,times}

\usepackage{amsmath,amsfonts,bm}

\def\eqref#1{equation~\ref{#1}}

\def\1{\bm{1}}

\DeclareMathAlphabet{\mathsfit}{\encodingdefault}{\sfdefault}{m}{sl}
\SetMathAlphabet{\mathsfit}{bold}{\encodingdefault}{\sfdefault}{bx}{n}

\usepackage{hyperref}
\usepackage{url}
\usepackage{graphicx}
\usepackage{booktabs}
\usepackage{multirow}
\usepackage{xcolor}
\usepackage{colortbl}
\usepackage{wrapfig}
\usepackage{caption}
\usepackage{makecell}
\usepackage{pifont}

\title{Breaking the SFT Plateau: Multimodal Structured Reinforcement Learning for Chart-to-Code Generation}

\author{Lei Chen \qquad Xuanle Zhao \qquad Zhixiong Zeng$^{\dag}$ \\
\textbf{Jing Huang \qquad Liming Zheng \qquad Yufeng Zhong \qquad Lin Ma$^{\ast}$} \\
Meituan \\
\texttt{zengzhixiong@meituan.com \qquad forest.linma@gmail.com} \\
Project: \url{https://github.com/DocTron-hub/MSRL}
}

\iclrfinalcopy 
\begin{document}
{\let\thefootnote\relax\footnotetext{$^{\dag}$ Project leader. $^{\ast}$ Corresponding author.}}


\maketitle

\begin{abstract}
While reinforcement learning (RL) has proven highly effective for general reasoning in vision-language models, its application to tasks requiring deep understanding of information-rich images and structured output generation remains underexplored. Chart-to-code generation exemplifies this challenge, demanding complex reasoning over visual charts to produce structured code. Supervised fine-tuning (SFT) alone is often insufficient, highlighting the need for effective RL strategies tailored to structured outputs.
In this paper, we systematically investigate the performance plateau of SFT through large-scale experiments and propose Multimodal Structured Reinforcement Learning (MSRL) for chart-to-code generation. We construct the largest training corpus to date, with 3 million chart-code pairs curated from real-world tables in arXiv papers, addressing the limitations of previous synthetic datasets. Despite achieving state-of-the-art performance, our experiments show that simply increasing SFT data eventually leads to diminishing improvements.
To break this plateau, MSRL employs a multi-granularity reward system that integrates both textual and visual feedback. At the textual level, rule-based rewards validate fine-grained code details, while at the visual level, a model-based reward assesses the structural similarity between rendered code and ground-truth charts. We implement a two-stage curriculum training strategy, first optimizing the model with textual rewards and then incorporating visual signals for further enhancement.
Experimental results demonstrate that MSRL substantially breaks the SFT plateau, improving high-level metrics by 6.2\% and 9.9\% on ChartMimic and ReachQA benchmarks, respectively. Notably, our method outperforms all existing approaches in the chart domain and achieves competitive results with advanced closed-source models.
\end{abstract}

\section{Introduction}
Large language models (LLMs) have demonstrated impressive reasoning capabilities on complex textual problems, including code generation and mathematical problem-solving \cite{openai2025o3o4mini, guo2025deepseek}. Pivotal to this success is the implementation of reinforcement learning (RL) to optimize the capacities of LLM to generate step-by-step reasoning \cite{wei2022chain} and verify final answers \cite{shao2024deepseekmath}. While multimodal large language models (MLLMs) have achieved impressive performance on established visual reasoning tasks like visual question answering \cite{zhang2024mm, huang2025vision}, their core reasoning capabilities remain limited, especially when tasks require processing information-dense images like charts \cite{xuchartmoe, chen2025chart}, and generating structured outputs like code \cite{xia2024chartx, zhao2025chartcoder}.

Among these challenges, chart-to-code generation~\cite{yangchartmimic, zhang2024gpt} stands out as a task of both high complexity and significant practical value. Charts are widely used to communicate scientific findings, business insights, and statistical data, yet their information is often locked in visual formats that are difficult for automated systems to interpret or repurpose. Enabling MLLMs to accurately convert charts into executable code not only advances their visual understanding and structured reasoning capabilities, but also opens up impactful applications, such as assisting researchers in automating the reproduction of scientific visualizations.
Although previous works have attempted to solve this task by using methods like Supervised Fine-Tuning (SFT) \cite{zhang2024tinychart, zhao2025chartcoder} or Direct Preference Optimization (DPO) \cite{rafailov2023direct, zhang2025enhancing} on synthetic data, their performance remains limited due to the simplistic patterns of the corpus and ineffective generalization strategies. Therefore, a promising direction is to leverage more realistic data and robust strategies to enhance model performance.

\begin{figure}[t]
    \centering
    \includegraphics[width=0.9\linewidth]{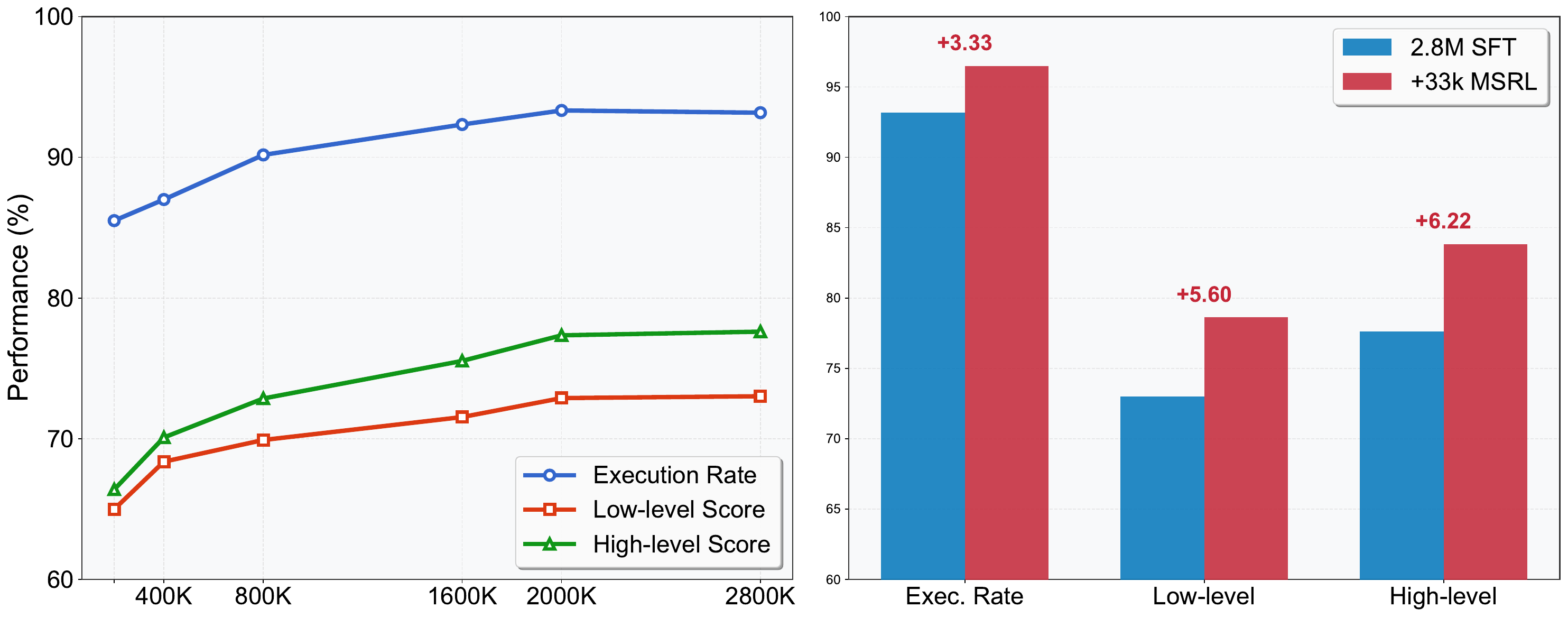}
    \vspace{-5pt}
    \caption{The SFT plateau and RL performance gain from our experiments. The left figure illustrates that scaling SFT data from 200k to 2.8M leads to a performance plateau after exceeding 2M data points. The right figure denotes the performance gain from our proposed MSRL training strategy.}
    \label{fig:sft_plateau}
    \vspace{-10pt}
\end{figure}

In this work, we first systematically investigate the SFT performance plateau through experiments at various data scales. To enable this analysis, we construct the largest training corpus to date for chart-to-code generation, covering 24 distinct chart types and comprising 3 million chart-code pairs. Our two-stage data pipeline curates high-quality tables from arXiv papers and then utilizes LLMs to generate diverse and complex plot codes, ensuring realistic distributions and avoiding monotonous trends.
Unlike prior works~\cite{ni2025point, chen2025learning} that apply RL directly after SFT, we establish saturated SFT baselines to isolate the true effectiveness of our proposed multimodal structured reinforcement learning (MSRL) strategy. Through experiments with varying SFT data scales, we demonstrate that simply increasing the volume of SFT data leads to a performance plateau that cannot be overcome by data quantity alone.
To break through this plateau, we propose the MSRL training strategy, which incorporates multi-granularity reward functions from both textual and visual aspects. For textual rewards, we standardize code formats in the training corpus and design a comprehensive reward function that verifies code correctness from five key perspectives. Recognizing that text-based rewards alone may overlook overall chart structure, we introduce a visual reward mechanism: generated code is rendered into images and an MLLM evaluates their similarity to the original input chart. By jointly leveraging these multi-granularity rewards, MSRL optimizes both global chart context and fine-grained code details.
We evaluate models on various chart-to-code benchmarks \cite{yangchartmimic, he2024distill}. The results demonstrate that it establishes a new state-of-the-art, as it outperforms all previous open-source models and rivals advanced proprietary MLLMs. In summary, the main contributions of this work are as follows:
\begin{itemize}
    \item We propose an MSRL training strategy that employs a multi-level reward function, which combines a rule-based component for assessing textual code correctness with a model-based evaluator for evaluating the visual fidelity of the rendered chart. Benefits from the MSRL training strategy, our proposed model achieves state-of-the-art (SOTA) performance compared to all other open-source MLLMs.
    \item We construct the first large-scale chart-to-code training corpus, generated using real-world tables as a data source. After applying several filtering methods, we curate a final dataset of 2.8 million samples for SFT and 33 thousand for RL.
    \item Through SFT experiments at various data scales, we identify the performance bottleneck, confirming that merely increasing SFT data quantity is insufficient to improve performance.
\end{itemize}

\section{Related Work}
\subsection{Chart MLLMs}
Chart understanding is a crucial research area encompassing both low-level tasks, such as data extraction \cite{liu2023deplot}, and high-level tasks like question answering \cite{masry2022chartqa} and summarization \cite{kantharaj2022chart}. Recent work has focused on training MLLMs on extensive, chart-specific datasets to enhance their understanding capabilities \cite{xia2024chartx, zhang2024tinychart,xia2023structchart}, leading to superior performance across various chart-related tasks. For example, ChartMoE \cite{xuchartmoe} proposes a mixture-of-experts (MoE) structure for multi-task pretraining. More recently, leveraging reinforcement learning~\cite{chen2025chart,AhmedMasry2025,rodriguez2025rendering} to enhance reasoning capacities has garnered significant interest. Both Chart-R1 and BigCharts-R1 optimize MLLMs using Chain-of-Thought (CoT) \cite{wei2022chain} reasoning data and reinforcement learning with verified rewards (RLVR) \cite{shao2024deepseekmath}.

\subsection{Reasoning MLLMs}
The success of large reasoning models (LRMs) like DeepSeek-R1 \cite{guo2025deepseek} spurs significant research into enhancing LLM reasoning through reinforcement learning with rule-based rewards \cite{shao2024deepseekmath}. This paradigm extends to the vision-language domain, where numerous works aim to improve the long-chain reasoning capabilities of MLLMs \cite{shen2025vlm, wang2025multimodal}. Following this trend, initial efforts such as Vision-R1 \cite{huang2025vision} and R1-OneVision \cite{yang2025r1} adapt the Group Relative Policy Optimization (GRPO) algorithm with multimodal reasoning data to enable long-form reasoning in VLMs. Subsequent works, including MMEureka \cite{meng2025mmeureka} and R1-Zero \cite{liu2025understanding}, further advance this area by introducing improved reinforcement learning strategies for visual long-term reasoning. Recently, approaches like Point-RFT \cite{ni2025point} utilize task-specific thinking frameworks to enhance reasoning capabilities further.

\subsection{MLLMs For Code}
In the rapidly advancing research area of multimodal code generation, foundational work has focused on establishing benchmarks to assess the capabilities of MLLMs. These benchmarks evaluate performance on tasks such as generating execution code for solving visual problems \cite{li2024mmcode, wang2025mathcoder,rodriguez2025starvector,rodriguez2024bigdocs,yang2024matplotagent} and HTML code from webpage screenshots \cite{si2024design2code, yun2024web2code, xiao2024interaction2code,awal2025webmmu}. 
Among these emerging tasks, chart-to-code generation has attracted significant interest due to the complexity of its visual inputs. This task challenges an MLLM to generate plotting code that accurately reproduces the chart image. 
Recently, several benchmarks have been proposed to evaluate MLLMs in this context, assessing a range of capabilities. For instance, benchmarks such as ChartMimic \cite{yangchartmimic}, Plot2Code \cite{wu2025plot2code}, and ChartX \cite{xia2024chartx} assess chart-to-code generation capabilities, while others like ChartEdit \cite{zhao2025chartedit} and $\text{ChartM}^\text{3}$ \cite{yang2025chartm} evaluate the editing functionalities.

\section{Methods}
\begin{figure}[t]
    \centering
    \includegraphics[width=0.95\linewidth]{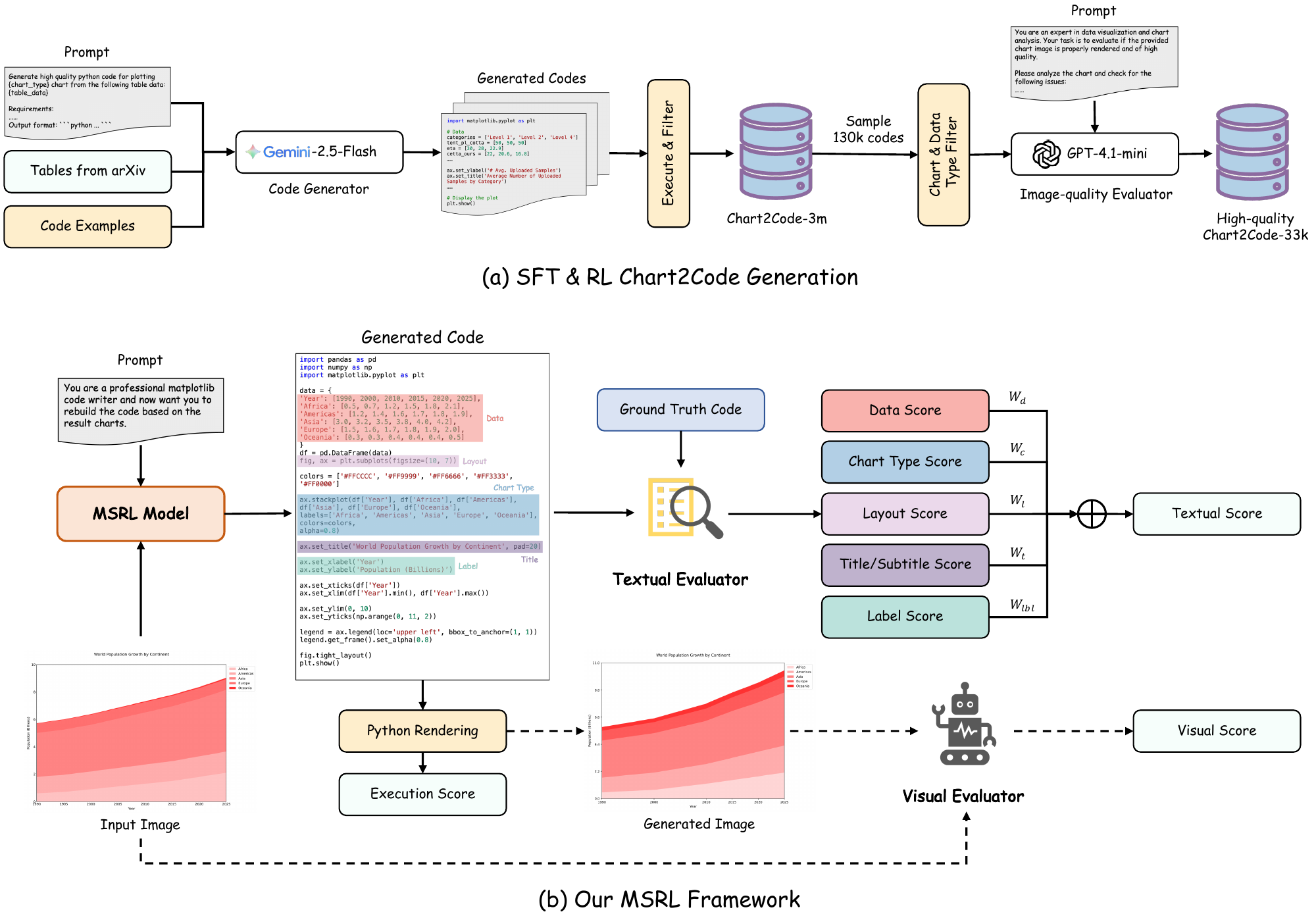}
    \caption{The data generation pipeline and our proposed MSRL framework. (a) Our pipeline prompts Gemini-2.5-Flash with tables from arXiv papers and example codes to generate plotting code. After execution and filtering, the chart-to-code dataset of 3M pairs is obtained. We then sample 130k of these pairs and apply three-stage filters (chart, data, vision) to curate a final, high-quality dataset of 33k examples for RL. (b) The framework of our proposed MSRL strategy. The textual reward is derived from a rule-based evaluation of the generated code across five distinct dimensions. An MLLM quantifies the visual reward based on the rendered image.} 
    \label{fig:pipeline}
    \vspace{-15pt}
\end{figure}

\subsection{Training Corpus Construction}
\label{sec:dataset}
While several chart-to-code generation datasets have been proposed, they are generally constrained by a reliance on purely synthetic data and limited scale, resulting in charts that exhibit simplistic trends and lack diversity. To overcome these challenges, we construct our dataset by crawling real-world tables from arXiv papers and then leveraging LLMs to generate the corresponding plotting code. 
To ensure separation from the ChartMimic benchmark, we exclusively use papers published in 2023 and earlier as our corpus.
We ensure data diversity and quality through several strategies, such as using high-quality seed code and filtering out non-executable code.

Through this process, we construct the largest training corpus to date, comprising 3 million chart-code pairs, to investigate the performance limits of SFT. However, previous research \cite{chen2025chart, huang2025vision} shows that separating the SFT and RL training corpora prevents overfitting to the SFT data format and enhances the exploration capacity of RL.
To curate a high-quality dataset for RL, we begin with a candidate corpus of 130 thousand samples from the overall training corpus. These candidate data samples are then subjected to a two-stage filtering strategy. The first filtering stage focuses on the code content, filtering based on chart and data type to ensure code diversity and that the output code is suitable for data extraction. Specifically, the chart types are filtered according to plot functions. To handle complex charts containing multiple chart types, we employ a tree-structured parsing method to identify each type present progressively. The data values are filtered according to the data definition format, retaining only data structured as one-dimensional arrays and non-nested dictionaries to facilitate extraction during the RL stage.
This step narrows the candidate corpus to 45 thousand samples. 
The second stage addresses visual quality, employing GPT-4.1-mini \cite{openai2024gpt41} as a visual LLM judge to filter for high-fidelity chart images. To mitigate potential bias from the MLLM judger, we manually inspect 100 samples that are assigned a score of 0 by the model, and observe over 90\% consistency with human judgment.
This process yields the final RL training corpus of 33 thousand high-quality samples. The remaining samples are utilized for SFT.
The comparisons of our dataset with relevant chart-to-code datasets are listed in Table~\ref{tab:existing_data}. Our dataset features a completely realistic data source, more challenging multi-chart content, and richer visual diversity enabled by more API types.

\begin{table}[h]
\setlength{\tabcolsep}{3pt}
\vspace{-10pt}
\small
\caption{Comparison of existing chart-to-code generation datasets. \ding{52}\rotatebox[origin=c]{-9.2}{\kern-0.7em\ding{55}} indicates that ChartMoE uses both data from real CSV tables and data generated by LLMs. "API types" refers to the number of different Matplotlib API types covered by the dataset.}
\label{tab:existing_data}
\centering
\begin{tabular}{l|c|c|c|c|c}
\toprule
\textbf{Dataset} & \textbf{\makecell{Realistic\\data source}} & \textbf{Multi-chart} & \textbf{Chart types} & \textbf{Data samples} & \textbf{API types} \\
\midrule
ChartLlama  & \ding{55}    & \ding{55}    & 10  & 11k   & 83  \\
ChartMoE    & \ding{52}\rotatebox[origin=c]{-9.2}{\kern-0.7em\ding{55}} & \ding{55}   & $<$20 & 800k  & -     \\
Chart2Code  & \ding{55}    & \ding{52}   & 15  & 3k    & 168  \\
ChartCoder  & \ding{55}    & \ding{55}    & 27  & 115k  & 187  \\
Ours        & \ding{52}   & \ding{52}   & 24  & 3M    & 1,555  \\
\bottomrule
\end{tabular}
\vspace{-15pt}
\end{table}

\subsection{Supervised Fine-tuning}
Firstly, we perform SFT on Qwen2.5-VL as the base model using our generated chart-to-code dataset. To explore the effectiveness of data scaling and identify the SFT plateau, we fine-tune our model on six distinct training corpus sizes, created by partitioning our main dataset into subsets of 200k, 400k, 800k, 1.6M, 2M and 2.8M samples. 
The model is trained by minimizing the standard autoregressive objective, which is the negative log-likelihood of the target sequence.
As demonstrated in Section~\ref{sec:analysis} and Figure~\ref{fig:sft_plateau}, merely scaling the SFT data volume ultimately results in a performance plateau. Increasing the SFT dataset from 2M to 2.8M samples results in no further performance gains.

\subsection{Multimodal Structured Reinforcement Learning}
Although SFT can significantly enhance the chart-to-code capabilities of MLLMs, our experiments reveal that this approach ultimately reaches a performance plateau, likely due to the limitation that SFT restricts the model's capacity to restore detailed information. This limitation stems from the core drawback of the SFT objective that it treats every token in the target sequence with uniform importance. Therefore, SFT is ineffective in optimizing plotting code generation, as the plotting code comprises largely of boilerplate snippets such as \texttt{plt.plot}, with critical information like specific data values or styling parameters appearing infrequently. Consequently, employing RL with a reward function designed to prioritize the accuracy of such critical content presents a promising approach for enhancing code generation capacity. To this end, we propose a two-stage RL training strategy utilizing multi-granularity reward functions that combine both textual and visual feedback.

\textbf{Textual Reward} While the RLVR method provides a powerful mechanism to guide models by measuring the correctness of detailed information, its practical application is challenged by the stylistic diversity of generated code, particularly in data definitions and function calls. This stylistic heterogeneity leads to difficulty in comprehensively extracting and parsing the critical information for reward computation. To mitigate this, we introduce a code normalization step that maps each generated output to a canonical representation before calculating the reward. This process ensures the textual reward function is invariant to syntactic variations, providing a more suitable method for RLVR. 
Specifically, the textual reward is a granular rule-based accuracy score, which is a weighted average derived from evaluating key aspects of the generated code. We assess data values using soft value matching, chart types with hard string matching, layout with hard value matching and elements like titles and labels via edit distance. The soft value matching allows a relative error tolerance of $\pm$5\%. Separately, we compute an execution reward, which is a binary score indicating whether the generated code can be executed successfully. These two rewards ensure that both the fidelity of critical information and the executability of the code are considered during RL optimization.

\textbf{Visual Reward} However, our analysis reveals that purely textual rewards generally focus on fine-grained details, while the overall structure of the rendered chart image is not considered. The chart-to-code task requires generating code that replicates the overall style of charts. To this end, we propose visual reward feedback~\cite{gu2025surveyllmasajudge}. Specifically, the generated code is executed to render a chart image, and we then utilize MLLMs to score the visual similarity between the generated chart and the input. This score is then normalized to serve as the visual reward.
For the visual reward, we employ Qwen2.5-VL-72B \cite{bai2025qwen25} as the evaluation model, which compares the chart image rendered by the generated code against the ground-truth chart across six key aspects: chart type, layout, text content, data, style, and clarity. The resulting scores are then normalized to form the final reward. Crucially, code that fails to render an image receives a reward of 0. The framework of our proposed MSRL is denoted in Figure~\ref{fig:pipeline}.

\textbf{Two-stage RL} 
Obtaining visual rewards through model evaluation requires substantial computational time. To ensure training efficiency, our two-stage RL strategy first optimizes the model using only a textual reward. In the second stage, we introduce a hybrid reward, which combines the textual reward with a visual one, to fine-tune the model for visual fidelity using a reduced number of samples. This approach allows us to balance computational cost and model performance, achieving strong visual quality without excessive resource consumption.
To properly balance the different training objectives, the total reward $R$ for a completed trajectory is calculated as a weighted sum of the following three components:
\begin{equation}
R = w_t R_{\text{text}} + w_v R_{\text{vis}} + w_e R_{\text{exec}}
\end{equation}
where $w_t$, $w_v$, and $w_e$ are hyperparameters that balance the contribution of each component. We adopt the Group Relative Policy Optimization (GRPO)~\cite{shao2024deepseekmath} algorithm for RL. GRPO optimizes policy by leveraging group-wise relative advantages among sampled responses, which is well-suited for our customized reward designs of charts.


\section{Experiment}
\subsection{Baselines and Benchmarks}
We evaluate MSRL against three categories of state-of-the-art models. The first category comprises general-domain, open-source Multimodal Large Language Models (MLLMs): InternVL2 (8B, 26B) \cite{chen2024internvl}, Qwen2-VL (7B, 72B) \cite{wang2024qwen2}, and Qwen2.5-VL-7B \cite{bai2025qwen25}. The second includes proprietary models: GeminiProVision \cite{team2023gemini}, Claude-3-opus \cite{anthropic2024claude}, GPT-4V \cite{openai2023gpt4v}, and GPT-4o \cite{openai2024gpt4o}. The third category consists of chart-specific MLLMs: ChartLlama \cite{han2023chartllama}, Tinychart \cite{zhang2024tinychart}, ChartVLM \cite{xia2024chartx}, Chart2Code \cite{zhang2025enhancing} and ChartCoder \cite{zhao2025chartcoder}.

All models were evaluated in a zero-shot setting on two distinct benchmarks: ChartMimic \cite{yangchartmimic} and ReachQA \cite{he2024distill}. For ReachQA, we adopt the evaluation setting of Chart2Code \cite{zhang2025enhancing}, using the 500 plotting scripts from its test set. For ChartMimic, we utilize the 600 examples from the latest version of the Direct Mimic task.

\begin{table}[t]
\setlength{\tabcolsep}{2pt}
\caption{Evaluation results of various models on the ChartMimic Direct Mimic and ReachQA benchmarks. $\ddag$ denotes the updated result with code\_pass constraints reported in its repository.
}
\vspace{-5pt}
\label{tab:main_results}
\centering
\begin{tabular}{l|r|ccc|ccc}
\toprule
\multirow{2}{*}{\textbf{Model}} & \multirow{2}{*}{\textbf{Params}} & \multicolumn{3}{c|}{\textbf{ChartMimic}} & \multicolumn{3}{c}{\textbf{ReachQA}} \\
 & & Exec.Rate & Low-Level & High-Level & Exec.Rate & Low-Level & High-Level\\ 
\midrule
\multicolumn{8}{c}{\it{Proprietary}} \\ 
\midrule
GeminiProVision & - & 68.2 & 53.8 & 53.3 & 74.0 & 67.0 & 67.8 \\
Claude-3-opus & - & 83.3 & 60.5 & 60.1 & 89.0 & 51.7 & 61.1 \\
GPT-4V & - & 91.2 & 76.4 & 78.9 & 88.0 & 69.5 & 78.6 \\
GPT-4o & - & 93.2 & \textbf{79.0} & 83.5 & 92.8 & 81.8 & 84.0 \\
\midrule
\multicolumn{8}{c}{\it{Open-Source General-Domain}} \\
\midrule
Qwen2-VL-7B & 7B & 67.0 & 32.9 & 35.0 & 55.4 & 22.6 & 29.3 \\
Qwen2.5-VL-7B & 7B & 73.2 & 44.6 & 41.6 & 62.2& 36.9 & 37.6 \\
InternVL2-8B & 8B & 61.8 & 34.4 & 38.9 & 50.8 & 24.1 & 24.2 \\
InternVL2-26B & 26B & 69.3 & 41.4 & 47.4 & 55.4 & 29.0 & 28.8 \\
Qwen2-VL-72B & 72B & 73.3 & 54.4 & 50.9 & 77.2 & 50.0 & 48.1 \\
\midrule
\multicolumn{8}{c}{\it{Open-Source Chart-Domain}} \\
\midrule
ChartLlama & 13B& 57.5 & 24.8 & 28.1 & 54.8 & 11.1 & 8.1 \\
TinyChart & 3B &  42.5 & 26.3 & 25.9 & 34.4 & 11.6 & 11.2 \\
ChartVLM-L & 14B & 19.5 & 15.8 & 13.9 & 8.2 & 2.1 & 3.9 \\
Chart2Code & 7B & 62.1 & 42.9 & 33.3 & 63.6 & 52.3 & 49.7 \\
ChartCoder & 7B & 91.4 & 72.5$^{\ddag}$ & 74.0 & 83.8 & 67.9 & 69.4 \\
\midrule
MSRL-SFT & 7B & 93.2 & 73.0 & 77.6 & 92.2 & 78.6 & 80.0 \\
MSRL  & 7B & \textbf{96.5} & 78.6 & \textbf{83.8} & \textbf{98.2} & \textbf{86.1} & \textbf{89.9} \\
\bottomrule 
\end{tabular}
\vspace{-10pt}
\end{table}

\begin{table}[t]
  \caption{Detailed results of low-level scores on the ChartMimic Direct Mimic and ReachQA benchmarks. The ChartCoder performance metrics have been corrected due to a flawed evaluation setup.}
  \vspace{-5pt}
  \label{tab:low_detailed_results} 
  \centering
  \begin{tabular}{l|r|cccc|cccc}
    \toprule
    \multirow{2}{*}{\textbf{Model}} & \multirow{2}{*}{\textbf{Params}} & \multicolumn{4}{c|}{\textbf{ChartMimic}} & \multicolumn{4}{c}{\textbf{ReachQA}} \\
    & & Text & Layout & Type & Color & Text & Layout & Type & Color \\
    \midrule
    GPT-4o & - & \textbf{81.5} & 89.8 & \textbf{77.3} & 67.2 & 84.4 & 91.2 & 81.3 & 70.5 \\
    \midrule
    Qwen2-VL-72B & 72B & 43.2 & 80.5 & 54.6 & 39.4 & 41.0 & 52.2 & 59.1 & 47.8 \\
    Qwen2.5-VL-7B & 7B & 37.7 & 65.6 & 42.6 & 32.4 & 29.6 & 40.9 & 44.2 & 32.8 \\
    \midrule
    ChartCoder$^{\ddag}$ & 7B & 65.9 & 85.4 & 72.3 & 66.4 & 58.2 & 80.0 & 70.4 & 63.2 \\
    \midrule
    MSRL & 7B & 78.0 & \textbf{93.0} & 75.3 & \textbf{68.2} & \textbf{87.5} & \textbf{96.6} & \textbf{83.1} & \textbf{77.2} \\
    \bottomrule
  \end{tabular}
\end{table}

\begin{figure}[t]
    \centering
    \includegraphics[width=0.9\linewidth]{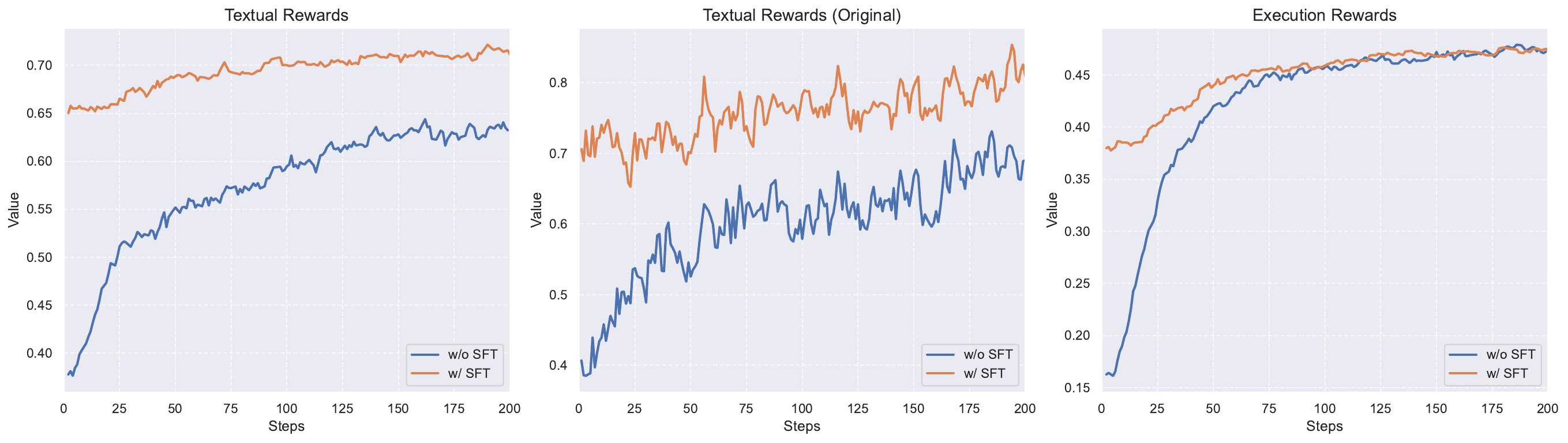}
    \vspace{-5pt}
    \caption{Comparison of textual reward and execution rate changes between baseline and SFT models during the RL stage.}
    \label{fig:rl_vs_sft_rl}
    \vspace{-15pt}
\end{figure}


\subsection{Main Results}
Our proposed MSRL establishes a new state-of-the-art among all open-source MLLMs, as detailed in Table~\ref{tab:main_results}. On the ChartMimic benchmark \cite{yangchartmimic}, it achieves an execution rate of 96.5, a low-level score of 78.6, and a high-level score of 83.8. Notably, its performance is not only superior to its open-source models but also comparable to that of GPT-4o, a significantly larger proprietary model. On the ChartMimic benchmark, MSRL outperforms ChartCoder, the previous leading Chart-domain MLLM specifically designed for chart-to-code generation. Unlike other works that yield suboptimal performances from fine-tuning open-source models, our initial MSRL-SFT model already establishes a new state-of-the-art among open-source models. However, MSRL-SFT still falls short of GPT-4o on both low-level and high-level performance metrics. Building upon this strong foundation, the MSRL strategy further boosts the performance, especially in high-level scores, demonstrating the effectiveness of our integrated approach. The performance gap between SFT and RL models is emphasized in Figure~\ref{fig:sft_plateau}.

We also conduct a detailed evaluation of low-level performance metrics, with the results presented in Table~\ref{tab:low_detailed_results}. Notably, MSRL outperforms all open-source models across low-level evaluation metrics and surpasses GPT-4o on the majority of evaluation metrics, which shows the strong capacity of MSRL in chart-to-code generation. As the GPT-4o is much larger than MSRL, the results demonstrate the effectiveness of our proposed methods. 

\label{sec:analysis}
\textbf{SFT Plateau} To investigate and demonstrate the effectiveness and efficiency of RL, we utilize various data scales for SFT to explore the upper limit of it. As illustrated in Figure~\ref{fig:sft_plateau}, the results reveal a clear trend, whereby model performance improves sharply as the data size increases to 400k samples, followed by a period of slower growth up to the 1.6M samples. Beyond this data scale, performance saturates and reaches a distinct plateau, showing negligible growth with any further increase in data scale. Based on this trend, even if the data is multiplied, the benefits brought by SFT will be minimal. Applying our proposed MSRL method yields significant performance gains across all metrics. On the ChartMimic benchmark, we observe an average improvement of 6\%, with notable increases in execution rate as well as in both low-level and high-level scores.

\begin{figure}[t]
    \centering
    \includegraphics[width=0.9\linewidth]{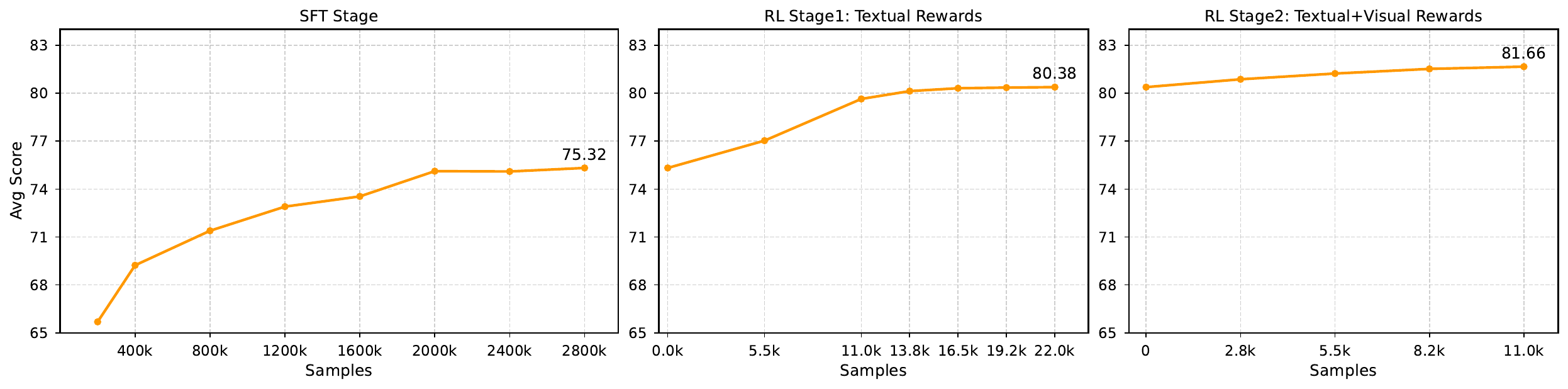}
    \vspace{-5pt}
    \caption{Performance curve of MSRL with increasing data scale. From left to right, the curves correspond to the SFT stage, RL stage 1, and RL stage 2. Avg Score denotes the average of the Low- and High-level scores in ChartMimic.}
    \label{fig:sft_rl_plateau}
    \vspace{-10pt}
\end{figure}

\textbf{RL Plateau} To investigate whether RL training exhibits a performance plateau, we present the performance curves of the two-stage RL training with the increase of data size, as shown in Figure~\ref{fig:sft_rl_plateau}. 
In the first stage using only textual rewards, the model breaks through the SFT plateau and achieves a significant performance improvement of approximately 5\%, converging at a maximum of 22k training samples.
In the second stage with multimodal rewards, the model's performance is further improved by about 1.5\%, and almost converges at a maximum of 11k training samples.
Our experiments effectively capture the performance plateau and improvements within a practical data size range, providing strong evidence for the effectiveness of our two-stage RL strategy.

\subsection{Ablation Study}

\begin{wraptable}{r}{0.52\textwidth}
\vspace{-15pt}
\setlength{\tabcolsep}{3pt}
\caption{Ablation studies on SFT and RL stages. For two RL experiments, only a textual reward is used.}
\label{tab:sft_rl_ablate}
\vspace{-5pt}
\centering
\begin{tabular}{cc|ccc}
\toprule
\multirow{2}{*}{\textbf{SFT}} & \multirow{2}{*}{\textbf{RL}} & \multicolumn{3}{c}{\textbf{ChartMimic}} \\
& & Exec.Rate & Low-Level & High-Level \\
\midrule
& & 73.16 & 44.58 & 41.55 \\
\checkmark & & 93.17 & 73.02 & 77.61 \\
& \checkmark & 93.83 & 65.60 & 62.29 \\
\checkmark & \checkmark & \textbf{97.00} & \textbf{78.06} & \textbf{82.69} \\
\bottomrule 
\end{tabular}
\vspace{-5pt}
\end{wraptable}
\textbf{SFT/RL Ablation} To understand the impact of different training settings, we conduct ablation studies on both the SFT and RL training stages. In this ablation study, we use only textual rewards to ensure the results are not influenced by our improved RL algorithm, allowing us to demonstrate the interplay between SFT and RL. Using Qwen2.5-VL-7B-Instruct as the baseline model, the results are presented in Table~\ref{tab:sft_rl_ablate}. As the RL-only model underperforms on low-level and high-level scores compared to the SFT-only approach, only the execution rate is slightly higher due to the inclusion of an execution-based reward. Furthermore, the significantly larger performance gain from RL on a suboptimal baseline, compared to an SFT model, indicates that this improvement is partly attributable to the unsaturated state of baseline models. The performance gain on a saturated SFT model demonstrates that our proposed multi-granularity reward function can surpass the upper limit of SFT, achieving significantly superior performance. Figure~\ref{fig:rl_vs_sft_rl} compares the reward and execution rate for the baseline and SFT models during the RL training, showing that applying RL to the SFT model achieves a much higher textual reward.

\begin{table}[h]
\setlength{\tabcolsep}{3pt}
\vspace{-5pt}
\caption{Comparison of single-stage and two-stage RL training with different reward strategies. “T” and “V” represent textual and visual reward, respectively. All training is conducted on H800 GPUs.}
\vspace{-5pt}
\label{tab:rl_strategy}
\centering
\begin{tabular}{c|c|ccc|c}
\toprule
\multirow{2}{*}{\textbf{Rewards}} & \multirow{2}{*}{\textbf{Samples}}     & \multicolumn{3}{c|}{\textbf{ChartMimic}} & \multirow{2}{*}{\textbf{GPU Hours}} \\
 & & Exec.Rate & Low-Level & High-Level &  \\
\midrule
-            & -          & 93.17           & 73.02     & 77.61      & -             \\
(T, -)       & (22k, -)   & 97.00           & 78.06     & 82.69      & 240           \\
(V, -)       & (22k, -)   & 97.37           & 79.13     & 84.01      & 1344           \\
(T, T + V)   & (22k, 11k) & 97.50           & 79.50     & 83.81      & 912           \\
(T, T + V)   & (22k, 5.5k)  & 96.50  & 78.62 & 83.83 & 576 \\
\bottomrule
\end{tabular}
\vspace{-5pt}
\end{table}

\textbf{Reward and RL Training Strategy Ablation} To thoroughly evaluate the effects of reward design and RL training strategies, we conduct ablation experiments comparing textual rewards, visual rewards, and their combinations. As shown in Table~\ref{tab:rl_strategy}, both reward types improve model performance, but differ notably in resource consumption and efficiency. Visual rewards provide notable enhancements in code details and visual fidelity, but require more computational resources due to the overhead of image rendering and model evaluation. In contrast, textual rewards achieve competitive performance with much lower resource requirements.  
To balance efficiency and effectiveness, we adopt a two-stage RL strategy.
The ablation results yield several important insights: 1) Both textual and visual rewards lead to significant improvements over the SFT model, with visual rewards being more effective for visual fidelity. 2) The high resource consumption of visual rewards highlights the importance of balancing performance and efficiency. 3) The two-stage RL strategy achieves the trade-off by first using low-cost textual rewards, followed by high-cost visual rewards with a limited number of samples.
Based on these findings, we empirically report the two-stage training strategy with 50\% of the training samples in our experiments, as it offers a better trade-off between performance and consumption efficiency.

\textbf{Generalization to Unseen Plotting Libraries} To evaluate whether MSRL can transfer its capabilities to other plotting libraries, we collect 107 Seaborn and 150 Plotly images from their official websites. Since we only use Matplotlib-style code to generate charts, the new test data are out-of-domain. We perform inference with both Qwen2.5-VL-7B and MSRL on these two small test sets, calculating execution rates and high-level scores according to the ChartMimic protocol. As shown in Table~\ref{tab:unseen_libs}, the results demonstrate that MSRL exhibits generalization to unseen Seaborn- and Plotly-style images.

\begin{table}[h]
\setlength{\tabcolsep}{3pt}
\vspace{-5pt}
\caption{Performance of Qwen2.5-VL-7B and MSRL on Seaborn and Plotly test sets.}
\label{tab:unseen_libs}
\centering
\vspace{-5pt}
\begin{tabular}{c|cc|cc}
\toprule
\multirow{2}{*}{\textbf{Model}} & \multicolumn{2}{c|}{\textbf{Seaborn}} & \multicolumn{2}{c}{\textbf{Plotly}} \\
 & Exec.Rate & High-Level & Exec.Rate & High-Level \\
\midrule
Qwen2.5-VL-7B & 69.2 & 25.7 & 62.7 & 22.6 \\
MSRL & \textbf{85.1} & \textbf{30.5} & \textbf{90.0} & \textbf{35.9} \\
\bottomrule 
\end{tabular}
\vspace{-5pt}
\end{table}

\begin{figure}[t]
    \centering
    \includegraphics[width=0.9\linewidth]{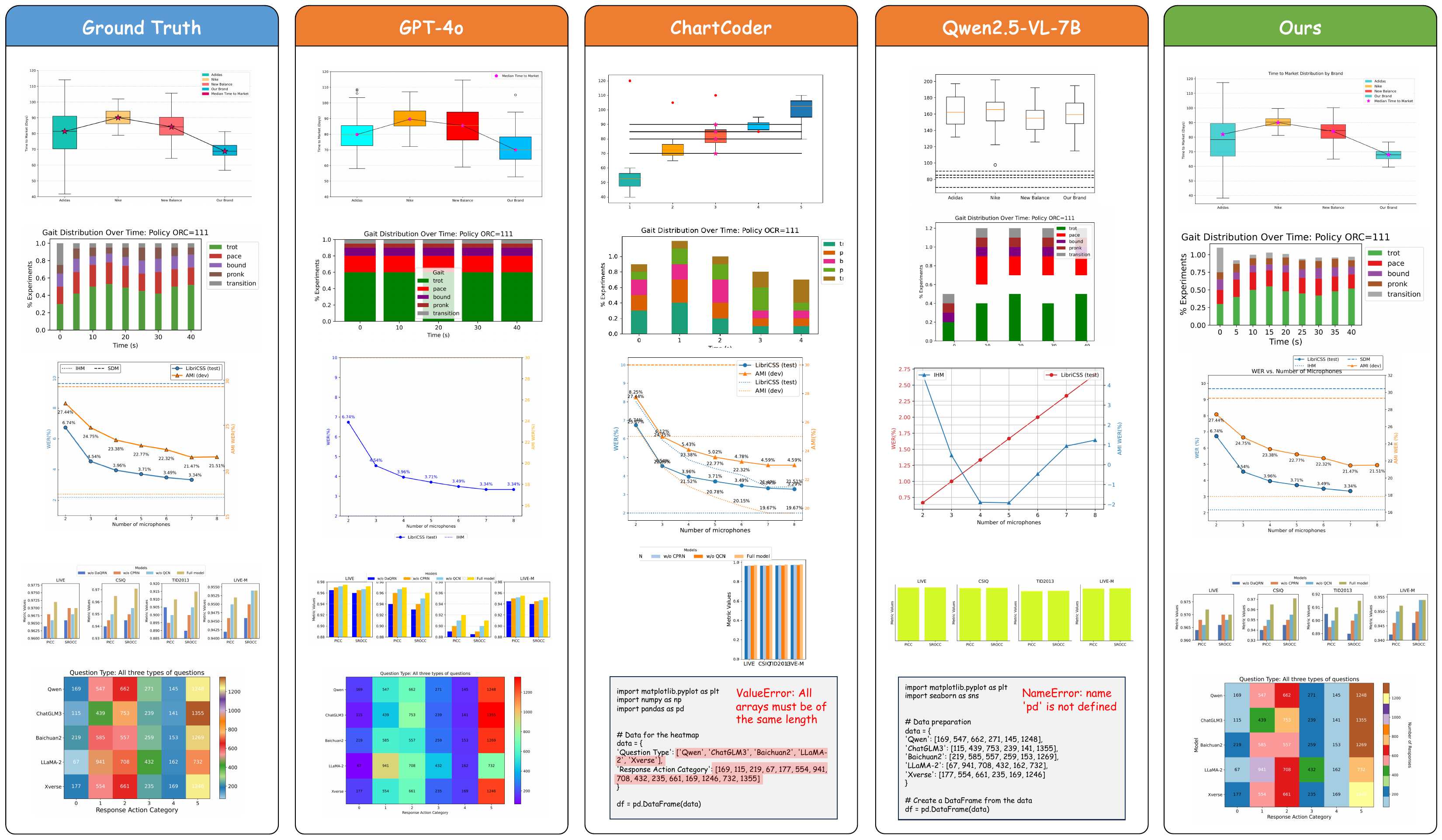}
    \vspace{-5pt}
    \caption{Showcasing charts generated by MSRL compared to proprietary and open-source MLLMs. The first column shows the ground-truth chart images. Columns 2–4 display charts rendered from the code generated by GPT-4o, ChartCoder, and Qwen2.5-VL-7B, while the last column shows charts rendered by our MSRL model.}
    \label{fig:output_comp}
    \vspace{-10pt}
\end{figure}

\subsection{Visualization}
To provide a qualitative perspective on performance, we showcase several illustrative examples. As visually demonstrated in Figure~\ref{fig:output_comp}, the charts generated by MSRL exhibit a markedly higher fidelity to the ground truth when compared with leading open-source MLLMs, even GPT-4o. In particular, the data values and layouts of charts generated by MSRL are significantly more precise.

\section{Conclusion}
In this work, we investigate the performance plateau of Supervised Fine-Tuning (SFT) in chart-to-code generation and propose a Multimodal Structured Reinforcement Learning (MSRL) strategy to overcome it. 
We begin by constructing a 3M-pair training corpus from real-world tables to systematically demonstrate that merely increasing SFT data leads to diminishing returns. 
To address this, we introduce the MSRL, which optimizes the saturated SFT model using a two-stage training process guided by a multi-granularity reward function that assesses both textual and visual correctness. 
Our approach achieves new state-of-the-art performance on standard benchmarks, surpassing all open-source models and rivaling powerful proprietary ones. We make significant contributions to the field, including the curation of a large-scale dataset for a definitive empirical analysis of the SFT performance bottleneck, and the development of the MSRL framework.

\section{Reproducibility Statement}
For datasets, we provide a detailed description of the data generation process in Section~\ref{sec:dataset} of the paper, along with dataset statistics and examples in Appendix~\ref{sec:data_detail}, and the exact prompts used for dataset construction in Appendix~\ref{sec:prompt}. For code implementation, we include the SFT and RL training code for MSRL in the supplementary materials, which contains complete training scripts and implementations of the multimodal reward functions.

\bibliography{iclr2026_conference}

@article{chen2024expanding,
  title={Expanding performance boundaries of open-source multimodal models with model, data, and test-time scaling},
  author={Chen, Zhe and Wang, Weiyun and Cao, Yue and Liu, Yangzhou and Gao, Zhangwei and Cui, Erfei and Zhu, Jinguo and Ye, Shenglong and Tian, Hao and Liu, Zhaoyang and others},
  journal={arXiv preprint arXiv:2412.05271},
  year={2024}
}

@article{han2023chartllama,
  title={Chartllama: A multimodal llm for chart understanding and generation},
  author={Han, Yucheng and Zhang, Chi and Chen, Xin and Yang, Xu and Wang, Zhibin and Yu, Gang and Fu, Bin and Zhang, Hanwang},
  journal={arXiv preprint arXiv:2311.16483},
  year={2023}
}

@misc{openai2025o3o4mini,
  author       = {{OpenAI}},
  title        = {Introducing OpenAI o3 and o4-mini},
  year         = {2025},
  month        = {April},
  day          = {16},
  howpublished = {\url{https://openai.com/index/introducing-o3-and-o4-mini/}},
  note         = {Accessed: 2025-07-14}
}

@article{ni2025point,
  title={Point-rft: Improving multimodal reasoning with visually grounded reinforcement finetuning},
  author={Ni, Minheng and Yang, Zhengyuan and Li, Linjie and Lin, Chung-Ching and Lin, Kevin and Zuo, Wangmeng and Wang, Lijuan},
  journal={arXiv preprint arXiv:2505.19702},
  year={2025}
}

@article{wang2025multimodal,
  title={Multimodal chain-of-thought reasoning: A comprehensive survey},
  author={Wang, Yaoting and Wu, Shengqiong and Zhang, Yuecheng and Yan, Shuicheng and Liu, Ziwei and Luo, Jiebo and Fei, Hao},
  journal={arXiv preprint arXiv:2503.12605},
  year={2025}
}

@inproceedings{zhao2025chartedit,
  title={Chartedit: How far are mllms from automating chart analysis? evaluating mllms’ capability via chart editing},
  author={Zhao, Xuanle and Liu, Xuexin and Haoyue, Yang and Luo, Xianzhen and Zeng, Fanhu and Li, Jianling and Shi, Qi and Chen, Chi},
  booktitle={Findings of the Association for Computational Linguistics: ACL 2025},
  pages={3616--3630},
  year={2025}
}

@inproceedings{chen2024internvl,
  title={Internvl: Scaling up vision foundation models and aligning for generic visual-linguistic tasks},
  author={Chen, Zhe and Wu, Jiannan and Wang, Wenhai and Su, Weijie and Chen, Guo and Xing, Sen and Zhong, Muyan and Zhang, Qinglong and Zhu, Xizhou and Lu, Lewei and others},
  booktitle={Proceedings of the IEEE/CVF Conference on Computer Vision and Pattern Recognition},
  pages={24185--24198},
  year={2024}
}

@article{wang2024qwen2,
  title={Qwen2-vl: Enhancing vision-language model's perception of the world at any resolution},
  author={Wang, Peng and Bai, Shuai and Tan, Sinan and Wang, Shijie and Fan, Zhihao and Bai, Jinze and Chen, Keqin and Liu, Xuejing and Wang, Jialin and Ge, Wenbin and others},
  journal={arXiv preprint arXiv:2409.12191},
  year={2024}
}

@article{team2023gemini,
  title={Gemini: a family of highly capable multimodal models},
  author={Team, Gemini and Anil, Rohan and Borgeaud, Sebastian and Alayrac, Jean-Baptiste and Yu, Jiahui and Soricut, Radu and Schalkwyk, Johan and Dai, Andrew M and Hauth, Anja and Millican, Katie and others},
  journal={arXiv preprint arXiv:2312.11805},
  year={2023}
}

@misc{anthropic2024claude,
  author       = {Anthropic},
  title        = {Introducing the Next Generation of Claude},
  year         = {2024},
  url          = {https://www.anthropic.com/news/claude-3-family}
}

@misc{openai2024gpt4o,
  author       = {OpenAI},
  title        = {GPT-4o},
  year         = {2024},
  url          = {https://openai.com/index/hello-gpt-4o},
  note         = {Accessed: 2024-05-13}
}

@misc{openai2023gpt4v,
  author       = {OpenAI},
  title        = {GPT-4V(ision) System Card},
  year         = {2023},
  url          = {https://openai.com/index/gpt-4v-system-card/}
}

@article{zhang2024tinychart,
  title={Tinychart: Efficient chart understanding with visual token merging and program-of-thoughts learning},
  author={Zhang, Liang and Hu, Anwen and Xu, Haiyang and Yan, Ming and Xu, Yichen and Jin, Qin and Zhang, Ji and Huang, Fei},
  journal={arXiv preprint arXiv:2404.16635},
  year={2024}
}

@article{xia2024chartx,
  title={Chartx \& chartvlm: A versatile benchmark and foundation model for complicated chart reasoning},
  author={Xia, Renqiu and Zhang, Bo and Ye, Hancheng and Yan, Xiangchao and Liu, Qi and Zhou, Hongbin and Chen, Zijun and Dou, Min and Shi, Botian and Yan, Junchi and others},
  journal={arXiv preprint arXiv:2402.12185},
  year={2024}
}

@article{wei2022chain,
  title={Chain-of-thought prompting elicits reasoning in large language models},
  author={Wei, Jason and Wang, Xuezhi and Schuurmans, Dale and Bosma, Maarten and Xia, Fei and Chi, Ed and Le, Quoc V and Zhou, Denny and others},
  journal={Advances in neural information processing systems},
  volume={35},
  pages={24824--24837},
  year={2022}
}

@article{shao2024deepseekmath,
  title={Deepseekmath: Pushing the limits of mathematical reasoning in open language models},
  author={Shao, Zhihong and Wang, Peiyi and Zhu, Qihao and Xu, Runxin and Song, Junxiao and Bi, Xiao and Zhang, Haowei and Zhang, Mingchuan and Li, YK and Wu, Y and others},
  journal={arXiv preprint arXiv:2402.03300},
  year={2024}
}

@inproceedings{li2024mmcode,
  title={MMCode: Benchmarking Multimodal Large Language Models for Code Generation with Visually Rich Programming Problems},
  author={Li, Kaixin and Tian, Yuchen and Hu, Qisheng and Luo, Ziyang and Huang, Zhiyong and Ma, Jing},
  booktitle={Findings of the Association for Computational Linguistics: EMNLP 2024},
  pages={736--783},
  year={2024}
}

@inproceedings{zhao2025chartcoder,
  title={ChartCoder: Advancing Multimodal Large Language Model for Chart-to-Code Generation},
  author={Zhao, Xuanle and Luo, Xianzhen and Shi, Qi and Chen, Chi and Wang, Shuo and Liu, Zhiyuan and Sun, Maosong},
  booktitle={Proceedings of the 63rd Annual Meeting of the Association for Computational Linguistics (ACL)},
  year={2025}
}

@article{huang2025vision,
  title={Vision-r1: Incentivizing reasoning capability in multimodal large language models},
  author={Huang, Wenxuan and Jia, Bohan and Zhai, Zijie and Cao, Shaosheng and Ye, Zheyu and Zhao, Fei and Xu, Zhe and Hu, Yao and Lin, Shaohui},
  journal={arXiv preprint arXiv:2503.06749},
  year={2025}
}

@article{liu2025understanding,
  title={Understanding r1-zero-like training: A critical perspective},
  author={Liu, Zichen and Chen, Changyu and Li, Wenjun and Qi, Penghui and Pang, Tianyu and Du, Chao and Lee, Wee Sun and Lin, Min},
  journal={arXiv preprint arXiv:2503.20783},
  year={2025}
}

@article{yang2025r1,
  title={R1-onevision: Advancing generalized multimodal reasoning through cross-modal formalization},
  author={Yang, Yi and He, Xiaoxuan and Pan, Hongkun and Jiang, Xiyan and Deng, Yan and Yang, Xingtao and Lu, Haoyu and Yin, Dacheng and Rao, Fengyun and Zhu, Minfeng and others},
  journal={arXiv preprint arXiv:2503.10615},
  year={2025}
}

@article{shen2025vlm,
  title={Vlm-r1: A stable and generalizable r1-style large vision-language model},
  author={Shen, Haozhan and Liu, Peng and Li, Jingcheng and Fang, Chunxin and Ma, Yibo and Liao, Jiajia and Shen, Qiaoli and Zhang, Zilun and Zhao, Kangjia and Zhang, Qianqian and others},
  journal={arXiv preprint arXiv:2504.07615},
  year={2025}
}

@article{guo2025deepseek,
  title={Deepseek-r1: Incentivizing reasoning capability in llms via reinforcement learning},
  author={Guo, Daya and Yang, Dejian and Zhang, Haowei and Song, Junxiao and Zhang, Ruoyu and Xu, Runxin and Zhu, Qihao and Ma, Shirong and Wang, Peiyi and Bi, Xiao and others},
  journal={arXiv preprint arXiv:2501.12948},
  year={2025}
}

@article{si2024design2code,
  title={Design2Code: How Far Are We From Automating Front-End Engineering?},
  author={Si, Chenglei and Zhang, Yanzhe and Yang, Zhengyuan and Liu, Ruibo and Yang, Diyi},
  journal={arXiv preprint arXiv:2403.03163},
  year={2024}
}

@inproceedings{zhang2024gpt,
  title={Is gpt-4v (ision) all you need for automating academic data visualization? exploring vision-language models’ capability in reproducing academic charts},
  author={Zhang, Zhehao and Ma, Weicheng and Vosoughi, Soroush},
  booktitle={Findings of the Association for Computational Linguistics: EMNLP 2024},
  pages={8271--8288},
  year={2024}
}

@article{bai2025qwen25,
  title={Qwen2. 5-vl technical report},
  author={Bai, Shuai and Chen, Keqin and Liu, Xuejing and Wang, Jialin and Ge, Wenbin and Song, Sibo and Dang, Kai and Wang, Peng and Wang, Shijie and Tang, Jun and others},
  journal={arXiv preprint arXiv:2502.13923},
  year={2025}
}

@inproceedings{yang2025chartm,
  title={ChartM3: Benchmarking Chart Editing with Multimodal Instructions},
  author={Yang, Donglu and Zhang, Liang and Yue, Zihao and Chen, Liangyu and Xu, Yichen and Wang, Wenxuan and Jin, Qin},
  booktitle={Proceedings of the 33rd ACM International Conference on Multimedia},
  pages={5001--5009},
  year={2025}
}

@article{chen2025chart,
  title={Chart-R1: Chain-of-Thought Supervision and Reinforcement for Advanced Chart Reasoner},
  author={Chen, Lei and Zhao, Xuanle and Zeng, Zhixiong and Huang, Jing and Zhong, Yufeng and Ma, Lin},
  journal={arXiv preprint arXiv:2507.15509},
  year={2025}
}

@article{he2024distill,
  title={Distill Visual Chart Reasoning Ability from LLMs to MLLMs},
  author={He, Wei and Xi, Zhiheng and Zhao, Wanxu and Fan, Xiaoran and Ding, Yiwen and Shan, Zifei and Gui, Tao and Zhang, Qi and Huang, Xuanjing},
  journal={arXiv preprint arXiv:2410.18798},
  year={2024}
}

@inproceedings{AhmedMasry2025,
  author = {Ahmed Masry and Abhay Puri and Masoud Hashemi and Juan A. Rodriguez and Megh Thakkar and Khyati Mahajan and Vikas Yadav and Sathwik Tejaswi Madhusudhan and Alexandre Piche and Dzmitry Bahdanau and Christopher Pal and David Vazquez and Enamul Hoque Prince  and Perouz Taslakian and Sai Rajeswar Mudumba and Spandana Gella},
  title = {BigCharts-R1: Enhanced Chart Reasoning with Visual Reinforcement Finetuning},
  booktitle = {Conference on Language Modeling (COLM)},
  year = {2025}
}

@article{rafailov2023direct,
  title={Direct preference optimization: Your language model is secretly a reward model},
  author={Rafailov, Rafael and Sharma, Archit and Mitchell, Eric and Manning, Christopher D and Ermon, Stefano and Finn, Chelsea},
  journal={Advances in neural information processing systems},
  volume={36},
  pages={53728--53741},
  year={2023}
}

@article{zhang2025enhancing,
  title={Enhancing chart-to-code generation in multimodal large language models via iterative dual preference learning},
  author={Zhang, Zhihan and Cao, Yixin and Liao, Lizi},
  journal={arXiv preprint arXiv:2504.02906},
  year={2025}
}

@article{chen2025learning,
  title={Learning Only with Images: Visual Reinforcement Learning with Reasoning, Rendering, and Visual Feedback},
  author={Chen, Yang and Shen, Yufan and Huang, Wenxuan and Zhou, Shen and Lin, Qunshu and Cai, Xinyu and Yu, Zhi and Shi, Botian and Qiao, Yu},
  journal={arXiv preprint arXiv:2507.20766},
  year={2025}
}

@article{xiao2024interaction2code,
  title={Interaction2Code: Benchmarking MLLM-based Interactive Webpage Code Generation from Interactive Prototyping},
  author={Xiao, Jingyu and Wan, Yuxuan and Huo, Yintong and Wang, Zixin and Xu, Xinyi and Wang, Wenxuan and Xu, Zhiyao and Wang, Yuhang and Lyu, Michael R},
  journal={arXiv preprint arXiv:2411.03292},
  year={2024}
}

@article{wang2025mathcoder,
  title={MathCoder-VL: Bridging Vision and Code for Enhanced Multimodal Mathematical Reasoning},
  author={Wang, Ke and Pan, Junting and Wei, Linda and Zhou, Aojun and Shi, Weikang and Lu, Zimu and Xiao, Han and Yang, Yunqiao and Ren, Houxing and Zhan, Mingjie and others},
  journal={arXiv preprint arXiv:2505.10557},
  year={2025}
}

@misc{openai2024gpt41,
  author = {{OpenAI}},
  title = {GPT-4.1},
  year = {2025},
  howpublished = {\url{https://openai.com/index/gpt-4-1/}},
}

@article{meng2025mmeureka,
      title={MM-Eureka: Exploring the Frontiers of Multimodal Reasoning with Rule-based Reinforcement Learning},
      author={Fanqing Meng and Lingxiao Du and Zongkai Liu and Zhixiang Zhou and Quanfeng Lu and Daocheng Fu and Tiancheng Han and Botian Shi and Wenhai Wang and Junjun He and Kaipeng Zhang and Ping Luo and Yu Qiao and Qiaosheng Zhang and Wenqi Shao},
      year={2025},
      journal={arXiv preprint arXiv:2503.07365},
}

@inproceedings{zheng2024llamafactory,
  title={LlamaFactory: Unified Efficient Fine-Tuning of 100+ Language Models},
  author={Yaowei Zheng and Richong Zhang and Junhao Zhang and Yanhan Ye and Zheyan Luo and Zhangchi Feng and Yongqiang Ma},
  booktitle={Proceedings of the 62nd Annual Meeting of the Association for Computational Linguistics (Volume 3: System Demonstrations)},
  address={Bangkok, Thailand},
  publisher={Association for Computational Linguistics},
  year={2024},
  url={http://arxiv.org/abs/2403.13372}
}

@article{oquab2023dinov2,
  title={Dinov2: Learning robust visual features without supervision},
  author={Oquab, Maxime and Darcet, Timoth{\'e}e and Moutakanni, Th{\'e}o and Vo, Huy and Szafraniec, Marc and Khalidov, Vasil and Fernandez, Pierre and Haziza, Daniel and Massa, Francisco and El-Nouby, Alaaeldin and others},
  journal={arXiv preprint arXiv:2304.07193},
  year={2023}
}

@inproceedings{rodriguez2025starvector,
  title={Starvector: Generating scalable vector graphics code from images and text},
  author={Rodriguez, Juan A and Puri, Abhay and Agarwal, Shubham and Laradji, Issam H and Rodriguez, Pau and Rajeswar, Sai and Vazquez, David and Pal, Christopher and Pedersoli, Marco},
  booktitle={Proceedings of the Computer Vision and Pattern Recognition Conference},
  pages={16175--16186},
  year={2025}
}

@article{xia2023structchart,
  title={Structchart: Perception, structuring, reasoning for visual chart understanding},
  author={Xia, Renqiu and Zhang, Bo and Peng, Haoyang and Ye, Hancheng and Yan, Xiangchao and Ye, Peng and Shi, Botian and Qiao, Yu and Yan, Junchi},
  journal={arXiv preprint arXiv:2309.11268},
  year={2023}
}

@inproceedings{rodriguez2024bigdocs,
  title={BigDocs: An Open Dataset for Training Multimodal Models on Document and Code Tasks},
  author={Rodriguez, Juan A and Jian, Xiangru and Panigrahi, Siba Smarak and Zhang, Tianyu and Feizi, Aarash and Puri, Abhay and Suresh, Akshay Kalkunte and Savard, Fran{\c{c}}ois and Masry, Ahmed and Nayak, Shravan and others},
  booktitle={The Thirteenth International Conference on Learning Representations},
  year={2024}
}

@inproceedings{awal2025webmmu,
  title={Webmmu: A benchmark for multimodal multilingual website understanding and code generation},
  author={Awal, Rabiul and Massoud, Mahsa and Feizi, Aarash and Li, Zichao and Wang, Suyuchen and Pal, Christopher and Agrawal, Aishwarya and Vazquez, David and Reddy, Siva and Rodriguez, Juan A and others},
  booktitle={Proceedings of the 2025 Conference on Empirical Methods in Natural Language Processing},
  pages={25129--25156},
  year={2025}
}

@inproceedings{rodriguez2025rendering,
  title={Rendering-Aware Reinforcement Learning for Vector Graphics Generation},
  author={Rodriguez, Juan A and Zhang, Haotian and Puri, Abhay and Feizi, Aarash and Pramanik, Rishav and Wichmann, Pascal and Mondal, Arnab and Samsami, Mohammad Reza and Awal, Rabiul and Taslakian, Perouz and others},
  booktitle={Advances in Neural Information Processing Systems (NeurIPS 2025)},
  year={2025}
}

@misc{gu2025surveyllmasajudge,
      title={A Survey on LLM-as-a-Judge}, 
      author={Jiawei Gu and Xuhui Jiang and Zhichao Shi and Hexiang Tan and Xuehao Zhai and Chengjin Xu and Wei Li and Yinghan Shen and Shengjie Ma and Honghao Liu and Saizhuo Wang and Kun Zhang and Yuanzhuo Wang and Wen Gao and Lionel Ni and Jian Guo},
      year={2025},
      eprint={2411.15594},
      archivePrefix={arXiv},
      primaryClass={cs.CL},
      url={https://arxiv.org/abs/2411.15594}, 
}

@inproceedings{kantharaj2022chart,
  title={Chart-to-text: A large-scale benchmark for chart summarization},
  author={Kantharaj, Shankar and Leong, Rixie Tiffany and Lin, Xiang and Masry, Ahmed and Thakkar, Megh and Hoque, Enamul and Joty, Shafiq},
  booktitle={Proceedings of the 60th Annual Meeting of the Association for Computational Linguistics (Volume 1: Long Papers)},
  pages={4005--4023},
  year={2022}
}

@inproceedings{liu2023deplot,
  title={DePlot: One-shot visual language reasoning by plot-to-table translation},
  author={Liu, Fangyu and Eisenschlos, Julian and Piccinno, Francesco and Krichene, Syrine and Pang, Chenxi and Lee, Kenton and Joshi, Mandar and Chen, Wenhu and Collier, Nigel and Altun, Yasemin},
  booktitle={Findings of the Association for Computational Linguistics: ACL 2023},
  pages={10381--10399},
  year={2023}
}

@inproceedings{masry2022chartqa,
  title={Chartqa: A benchmark for question answering about charts with visual and logical reasoning},
  author={Masry, Ahmed and Do, Xuan Long and Tan, Jia Qing and Joty, Shafiq and Hoque, Enamul},
  booktitle={Findings of the association for computational linguistics: ACL 2022},
  pages={2263--2279},
  year={2022}
}

@inproceedings{yangchartmimic,
  title={ChartMimic: Evaluating LMM's Cross-Modal Reasoning Capability via Chart-to-Code Generation},
  author={Yang, Cheng and Shi, Chufan and Liu, Yaxin and Shui, Bo and Wang, Junjie and Jing, Mohan and XU, Linran and Zhu, Xinyu and Li, Siheng and Zhang, Yuxiang and others},
  booktitle={The Thirteenth International Conference on Learning Representations},
  year={2024}
}

@inproceedings{wu2025plot2code,
  title={Plot2code: A comprehensive benchmark for evaluating multi-modal large language models in code generation from scientific plots},
  author={Wu, Chengyue and Liang, Zhixuan and Ge, Yixiao and Guo, Qiushan and Lu, Zeyu and Wang, Jiahao and Shan, Ying and Luo, Ping},
  booktitle={Findings of the Association for Computational Linguistics: NAACL 2025},
  pages={3006--3028},
  year={2025}
}

@inproceedings{xuchartmoe,
  title={ChartMoE: Mixture of Diversely Aligned Expert Connector for Chart Understanding},
  author={Xu, Zhengzhuo and Qu, Bowen and Qi, Yiyan and Du, SiNan and Xu, Chengjin and Yuan, Chun and Guo, Jian},
  booktitle={The Thirteenth International Conference on Learning Representations},
  year={2024}
}

@inproceedings{yang2024matplotagent,
  title={MatPlotAgent: Method and Evaluation for LLM-Based Agentic Scientific Data Visualization},
  author={Yang, Zhiyu and Zhou, Zihan and Wang, Shuo and Cong, Xin and Han, Xu and Yan, Yukun and Liu, Zhenghao and Tan, Zhixing and Liu, Pengyuan and Yu, Dong and others},
  booktitle={Findings of the Association for Computational Linguistics ACL 2024},
  pages={11789--11804},
  year={2024}
}

@article{yun2024web2code,
  title={Web2code: A large-scale webpage-to-code dataset and evaluation framework for multimodal llms},
  author={Yun, Sukmin and Thushara, Rusiru and Bhat, Mohammad and Wang, Yongxin and Deng, Mingkai and Wang, Jinhong and Tao, Tianhua and Li, Junbo and Li, Haonan and Nakov, Preslav and others},
  journal={Advances in neural information processing systems},
  volume={37},
  pages={112134--112157},
  year={2024}
}

@inproceedings{zhang2024mm,
  title={MM-LLMs: Recent Advances in MultiModal Large Language Models},
  author={Zhang, Duzhen and Yu, Yahan and Dong, Jiahua and Li, Chenxing and Su, Dan and Chu, Chenhui and Yu, Dong},
  booktitle={Findings of the Association for Computational Linguistics ACL 2024},
  pages={12401--12430},
  year={2024}
}
\bibliographystyle{iclr2026_conference}

\appendix
\renewcommand{\thefigure}{\Alph{figure}}
\renewcommand{\thetable}{\Alph{table}}
\setcounter{figure}{0}
\setcounter{table}{0}

\section{Chart2Code Details}
\label{sec:data_detail}

\begin{table}[h]
\setlength{\tabcolsep}{3pt}
\vspace{-5pt}
\caption{The detailed chart types and corresponding quantities used in the SFT and RL phases.}
\label{tab:chart_types}
  \centering
  \footnotesize
  \begin{tabular}{c|ccccccccc}
    \toprule
    Split & Bar & Line & ErrorBar & Heatmap & Box & Scatter & Histogram & Radar & 3D \\
    \midrule
    SFT & 131,623 & 131,596 & 130,818 & 123,795 & 125,541 & 127,612 & 132,180 & 124,160 & 128,744 \\
    RL & 2,669 & 2,743 & 1,067 & 1,984 & 756 & 2,639 & 1,510 & 776 & 1,699 \\
    \midrule
    Split & Pie & ErrorPoint & Violin & Area & Bubble & Multi-axes & Ring & Rose & Treemap \\
    \midrule    
    SFT & 125,812 & 123,774 & 120,180 & 120,038 & 127,207 & 116,438 & 105,912 & 118,047 & 119,176 \\
    RL & 1,446 & 2,277 & 900 & 1,558 & 1,093 & 1,691 & 960 & 674 & 518 \\
    \midrule
    Split & Bar\_num & Contour & Density & Quiver & Graph & Funnel & Total \\
    \midrule
    SFT & 128,333 & 123,461 & 130,769 & 128,987 & 108,150 & 1,152 & 2,853,505 \\
    RL & 2,757 & 1,040 & 1,899 & 1,700 & - & - & 34,356\\
    \bottomrule
  \end{tabular}
\end{table}

\begin{figure}[t]
    \centering
    \includegraphics[width=0.7\linewidth]{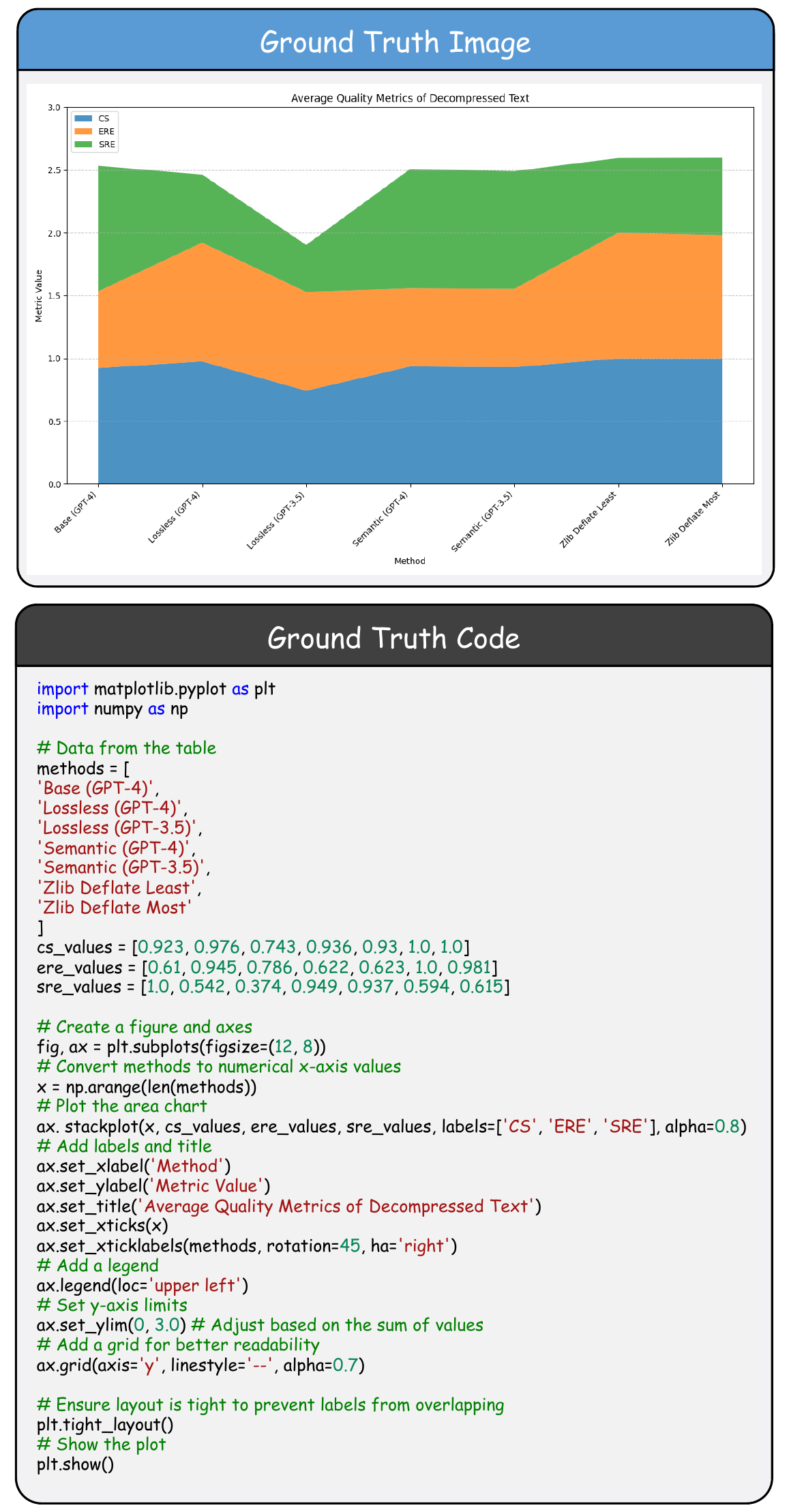}
    \vspace{-5pt}
    \caption{A case of chart-code pairs for area charts in our Chart2Code dataset.}
    \label{fig:data_show_1}
    \vspace{-5pt}
\end{figure}

\begin{figure}[t]
    \centering
    \includegraphics[width=0.7\linewidth]{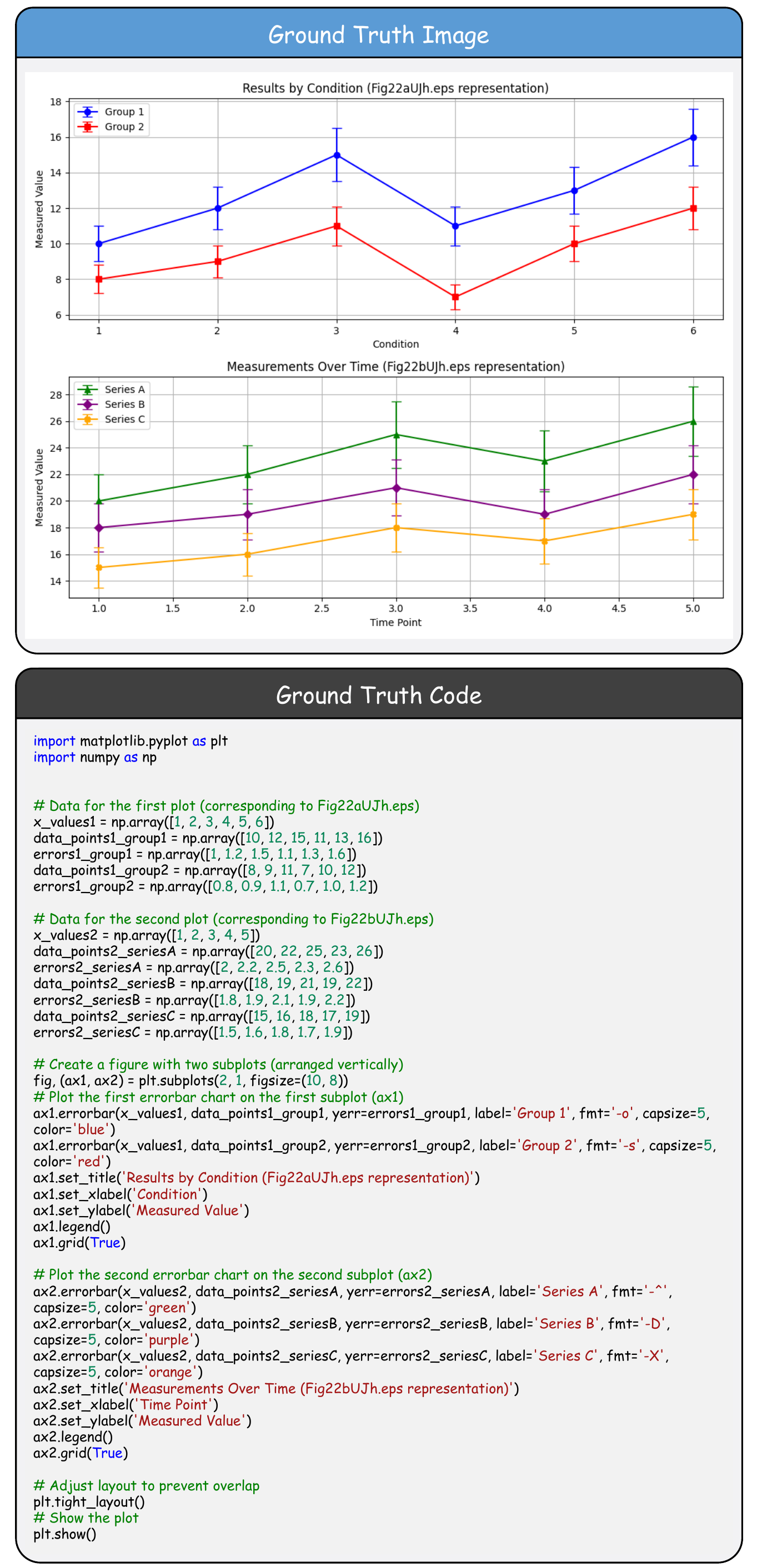}
    \vspace{-5pt}
    \caption{A case of chart-code pairs for errorbar charts in our Chart2Code dataset.}
    \label{fig:data_show_2}
    \vspace{-5pt}
\end{figure}

\begin{figure}[t]
    \centering
    \vspace{-5pt}
    \includegraphics[width=0.7\linewidth]{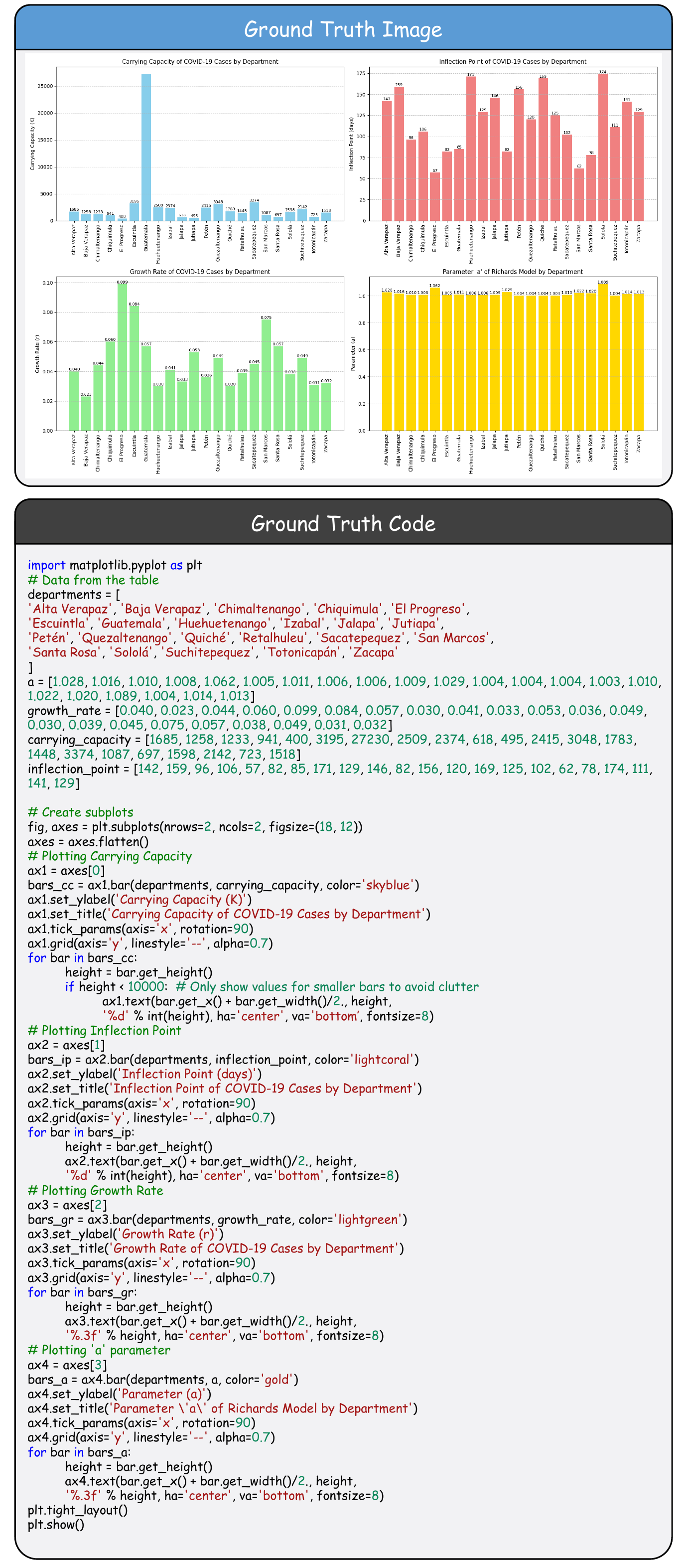}
    \vspace{-5pt}
    \caption{A case of chart-code pairs for bar\_num charts in our Chart2Code dataset.}
    \label{fig:data_show_3}
    \vspace{-5pt}
\end{figure}

We count the quantity and distribution of 24 chart types used in the SFT and RL stages respectively, as detailed in Table~\ref{tab:chart_types}. The distribution among the various types is relatively balanced. For the RL data, we filtered out two chart types, Graph and Funnel, which are not suitable for textual reward calculation. Furthermore, we present three visualization examples from the Chart2Code dataset, with increasing data density and complexity, as shown in Figures~\ref{fig:data_show_1}, \ref{fig:data_show_2} and \ref{fig:data_show_3}.

\section{Implementation Details}
We utilize 2.8M chart2code data for SFT and 33k curated high-quality chart2code data for RL. MSRL-SFT employs Qwen2.5VL-7B-Instruct as the baseline model and LLaMA-Factory~\cite{zheng2024llamafactory} for SFT. The SFT stage is trained for one epoch with a learning rate of 1e-5 and a batch size of 32. The training process takes 60 hours on 16 H800 GPUs. Built upon the SFT model, MSRL undergoes a two-stage RL process using the MM-EUREKA framework~\cite{meng2025mmeureka}. MSRL consists of two RL stages. Stage one utilizes 22k data pairs and a textual-only reward ($w_t=1$, $w_v=0$, $w_e=0.5$).  Stage two uses the remaining 11k pairs with a hybrid reward ($w_t=0.5$, $w_v=0.5$, $w_e=0.5$). The textual reward components are weighted as $w_d=0.4$, $w_c=0.3$, $w_l=0.1$, $w_t=0.1$, $w_{lbl}=0.1$. We deploy Qwen2.5-VL-72B-Instruct as the evaluation model with max\_model\_len set to 12,288 for calculating visual rewards.
For both RL stages, we train for one episode with a learning rate of 1e-6 and a batch size of 128. We set the generation temperature to 1, generate 8 rollouts per sample, and omit the KL divergence from the loss computation. The first and second stages required 10 and 24 hours of training, respectively, on 24 H800 GPUs.

\section{More Ablation Studies}
\label{sec:more_studies}
\begin{figure}[t]
    \centering
    \includegraphics[width=1.0\linewidth]{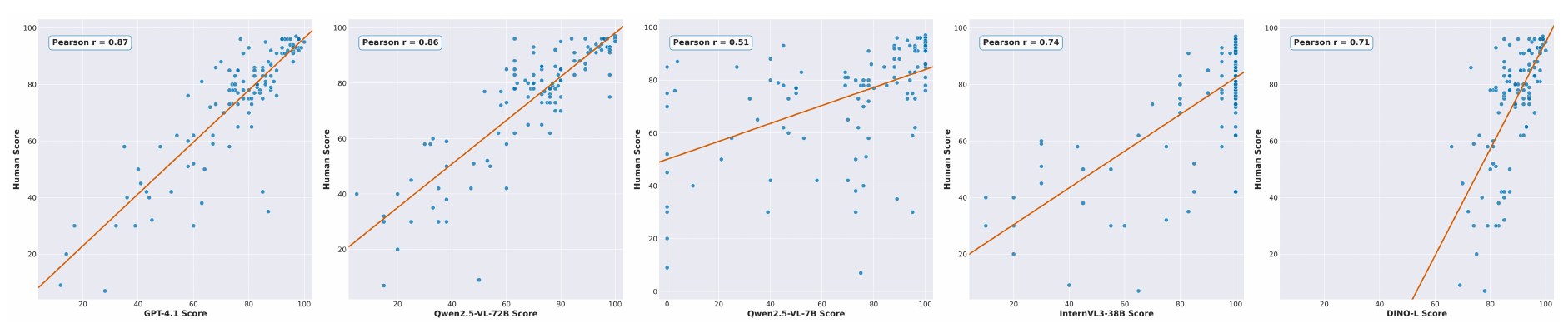}
    \vspace{-10pt}
    \caption{Scatter plots of model scores versus human gold scores for five visual evaluators (GPT-4.1, Qwen2.5-VL-72B, Qwen2.5-VL-7B, InternVL3-38B, and DINO-L). Pearson $r$ denotes the Pearson correlation coefficient between model and human scores.}
    \label{fig:diff_model_score}
    \vspace{-5pt}
\end{figure}

\textbf{The Use of Visual Evaluator}
To verify that the improvement in performance results from effective visual reward design rather than overfitting to the preferences of the VLLM judger, we conduct an ablation experiment to observe the distribution of sample scores when the visual evaluator changes.  Specifically, we randomly sample 200 instances from the RL data and use the MSRL-SFT model to perform chart-to-code prediction, rendering the resulting Python code into images. Following the ChartMimic scoring guidelines, human ratings of rendered samples serve as the gold scores. Next, we use GPT-4.1, Qwen2.5-VL-72B, Qwen2.5-VL-7B, and InternVL3-38B to score the images. Additionally, we compute the visual similarity between the ground-truth and rendered images using DINO-L. We calculate the Pearson correlation coefficients between the scores from these models and the human scores, and plot the distributions of model scores versus human scores. As shown in the Figure~\ref{fig:diff_model_score}, GPT-4.1 achieves the highest correlation with human judgment and has the most similar score distribution, followed by Qwen2.5-VL-72B, while Qwen2.5-VL-7B performs the worst. Considering both performance and budget, we select Qwen2.5-VL-72B as the visual evaluator, as it fairly reflects model capability and prevents reward hacking.

In addition, we introduce two vision models, InternViT~\cite{chen2024expanding} and DINOv2-L~\cite{oquab2023dinov2}, as evaluators for visual rewards, calculating the visual similarity between ground-truth and generated image features. As shown in Table~\ref{tab:visual_evaluator}, these vision models yield improved performance, but there remains a substantial gap compared to models employing VLMs as visual evaluators. This indicates that vision models trained on image-text matching assess similarity at a coarse level and do not adequately capture fine-grained differences in chart elements. Compared to Qwen2.5-VL-72B, smaller vision models such as DINO-L exhibit greater discrepancies in score distributions relative to human judgment.

\begin{table}[h]
\setlength{\tabcolsep}{3pt}
\caption{The ablation study of visual evaluators. "-" denotes the SFT model. All experiments exclude textual rewards.} 
\label{tab:visual_evaluator}
\vspace{-5pt}
\centering
\begin{tabular}{c|cc}
\toprule
\multirow{2}{*}{\textbf{Visual Evaluator}} & \multicolumn{2}{c}{\textbf{ChartMimic}} \\
& Low-Level & High-Level \\
\midrule
- & 73.0 & 77.6 \\
InternViT & 75.1 & 78.1 \\
DINOv2-L & 75.5 & 79.8 \\
Qwen2.5-VL-72B  & \textbf{77.8} & \textbf{81.9} \\
\bottomrule 
\end{tabular}
\vspace{-5pt}
\end{table}

\textbf{GRPO vs DPO} Intuitively, GRPO is naturally more suitable for chart-to-code tasks than DPO. The degree to which code restores a chart can be quantitatively measured by a set of well-defined rules, making it ideal for designing the reward function in GRPO. In contrast, DPO learns from preference pairs, and it is challenging to clearly define what constitutes "bad" code in chart-to-code tasks, which can lead to ambiguity in the model’s optimization direction. To ensure a thorough evaluation, we conduct an ablation study comparing GRPO and DPO. Since DPO requires the construction of positive and negative sample pairs, we follow the Chart2Code~\cite{zhang2025enhancing} and randomly perturb 2-3 attributes of the chart code to generate negative samples using GPT-4.1, resulting in an RL dataset of comparable size to GRPO for DPO training on the SFT model. As shown in Table~\ref{tab:dpo}, the performance improvement of DPO is significantly less than that of GRPO. This indicates both the superiority of the GRPO algorithm and the effectiveness of our multimodal structured reward design and two-stage curriculum learning.


\begin{table}[h]
\setlength{\tabcolsep}{3pt}
\caption{The ablation study of RL optimizers. "-" denotes the SFT model.}
\label{tab:dpo}
\vspace{-5pt}
\centering
\begin{tabular}{c|cc}
\toprule
\multirow{2}{*}{\textbf{RL Optimizer}} & \multicolumn{2}{c}{\textbf{ChartMimic}} \\
& Low-Level & High-Level \\
\midrule
-  & 73.0 & 77.6 \\
DPO & 74.0 & 78.9 \\
GRPO & \textbf{78.6} & \textbf{83.8} \\
\bottomrule 
\end{tabular}
\vspace{-5pt}
\end{table}

\textbf{Comparison with Existing Datasets} To highlight the value of our synthetic data, we compare the results of using our data and open-source data (i.e., ChartCoder) in both the SFT and RL stages, as shown in Table~\ref{tab:existing_comp}. The results show that models trained with our data achieve better performance in both stages. On the one hand, our sufficient data enable SFT to reach its full potential. On the other hand, our charts are generated from real tabular data, featuring more challenging multi-chart content and greater visual diversity, which are key factors contributing to the effectiveness of our dataset.

\begin{table}[h]
\setlength{\tabcolsep}{3pt}
\caption{Performance comparison with the ChartCoder dataset. We use Qwen2.5-VL-7B and MSRL-SFT as base models for SFT and RL for fine-tuning on the two datasets, respectively.}
\label{tab:existing_comp}
\vspace{-5pt}
\centering
\begin{tabular}{c|c|cc}
\toprule
\multirow{2}{*}{\textbf{Stage}} & \multirow{2}{*}{\textbf{Dataset}} & \multicolumn{2}{c}{\textbf{ChartMimic}} \\
& & Low-Level & High-Level \\
\midrule
\multirow{2}{*}{SFT} & ChartCoder & 68.5 & 68.0 \\
 & Ours & \textbf{73.0} & \textbf{77.6} \\
\midrule
\multirow{2}{*}{RL} & ChartCoder & 76.5 & 81.2 \\
& Ours & \textbf{78.6} & \textbf{83.8} \\
\bottomrule 
\end{tabular}
\vspace{-5pt}
\end{table}

\section{Qualitative Analysis}
\begin{figure}[t]
    \centering
    \includegraphics[width=1.0\linewidth]{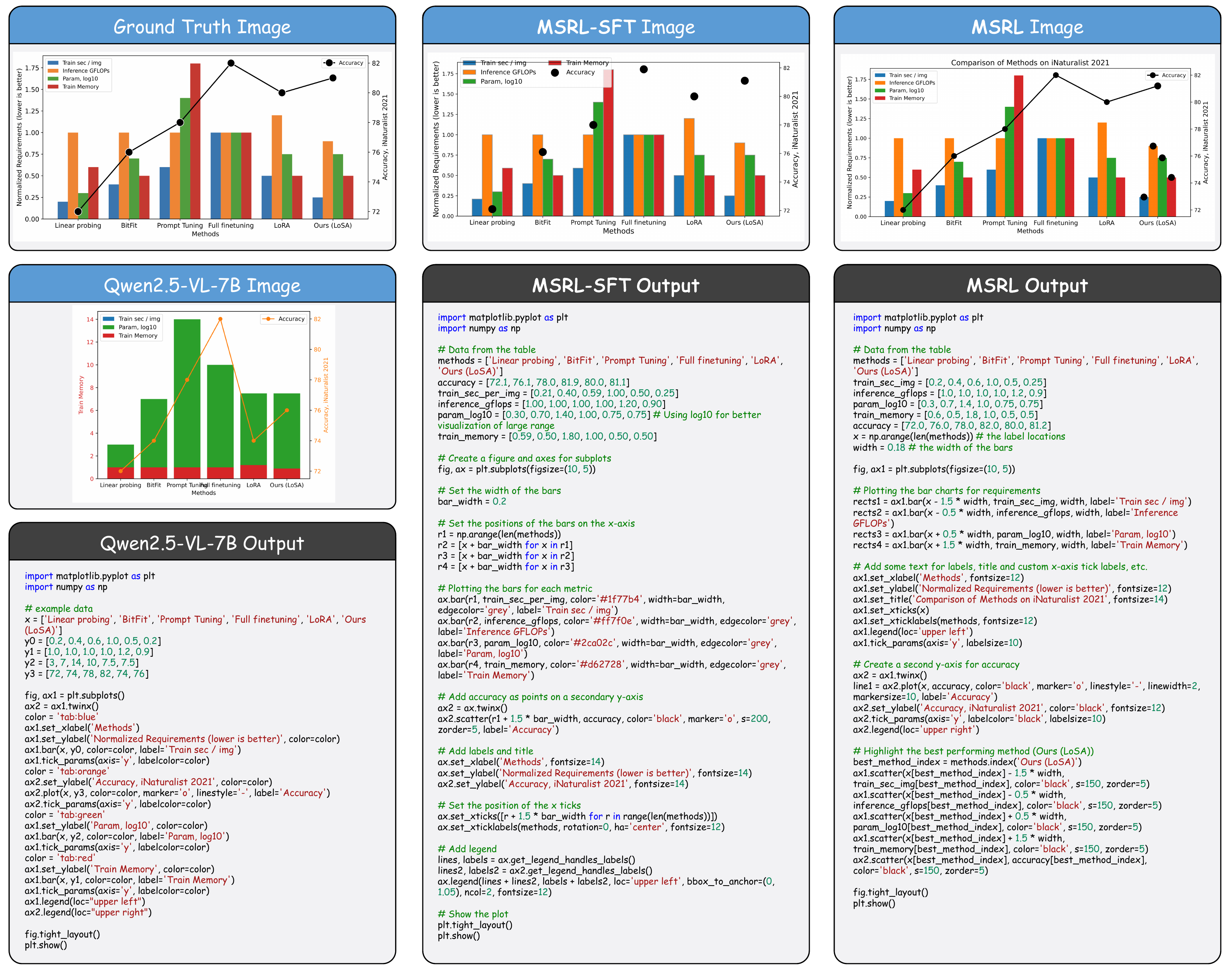}
    \caption{An example of comparing the code generated by different models with rendered images.}
    \label{fig:quantitative_analysis}
    \vspace{-5pt}
\end{figure}

\begin{figure}[t]
    \centering
    \includegraphics[width=1.0\linewidth]{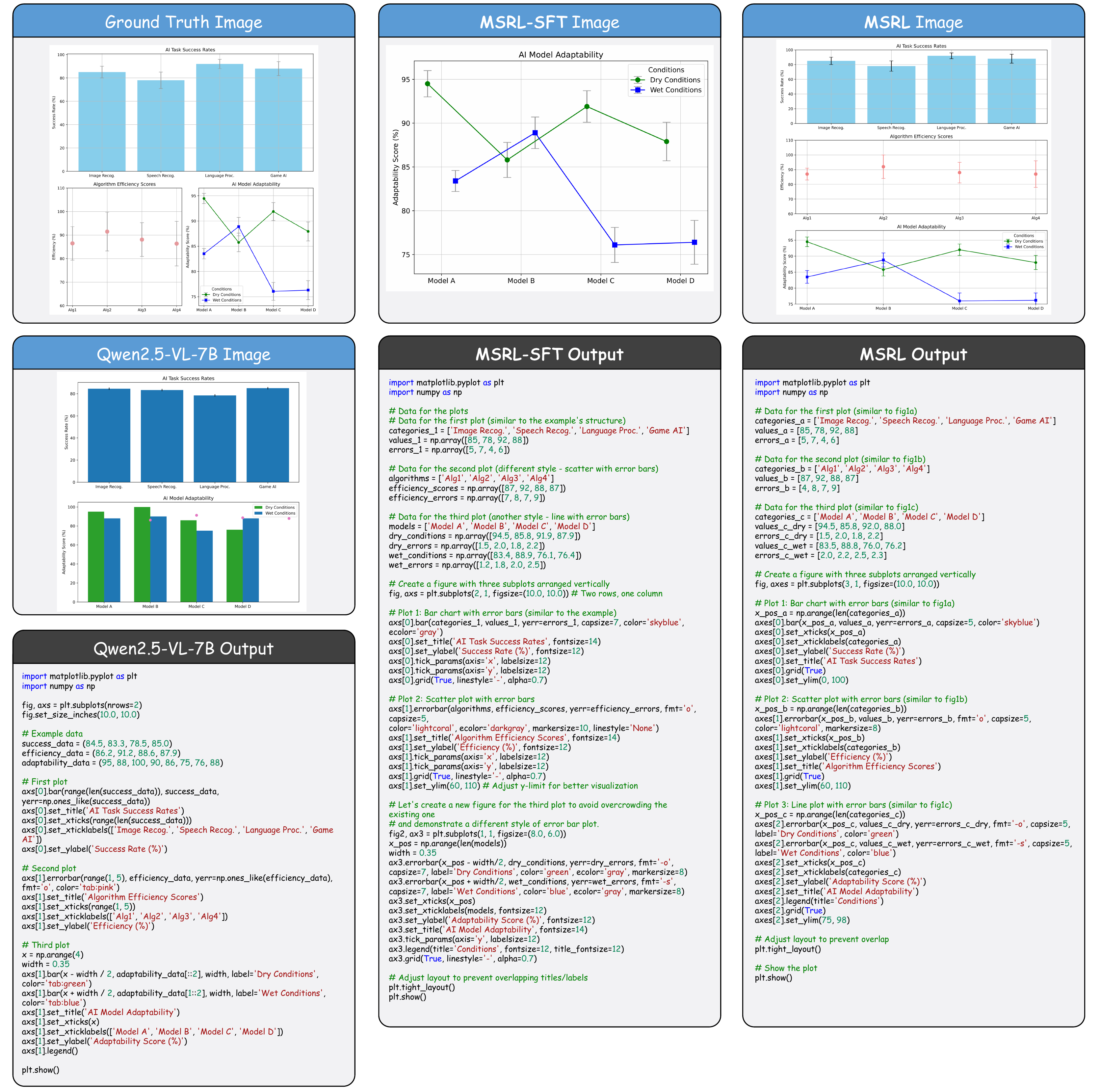}
    \caption{An example of comparing the code generated by different models with rendered images.}
    \label{fig:quantitative_analysis_2}
    \vspace{-5pt}
\end{figure}

Figures~\ref{fig:quantitative_analysis} and ~\ref{fig:quantitative_analysis_2} show two examples of generated code from different models and their corresponding rendered images. Compared to the initial Qwen2.5-VL-7B model, both MSRL-SFT and MSRL show significant improvements in chart type recognition, numerical extraction, and color identification. Compared to MSRL-SFT, MSRL performs better in chart type recognition and clarity. In Figure~\ref{fig:quantitative_analysis}, MSRL-SFT incorrectly identifies the line chart as a scatter plot, displays incorrect legends, and has overlapping issues with image boundaries. In Figure~\ref{fig:quantitative_analysis_2}, MSRL-SFT has layout syntax errors, where only the last subplot is successfully displayed out of three subplots. MSRL's results are more consistent with the original images, though it still has some minor issues in detail. For example, in Figure~\ref{fig:quantitative_analysis}, MSRL arbitrarily adds a title and scatter data points to the subcategory ``Ours (LoSA)". In Figure~\ref{fig:quantitative_analysis_2}, while MSRL successfully renders three subplots, the layout recognition is not sufficiently accurate, leaving room for further improvement.

\section{Prompt Template}
\label{sec:prompt}
To enhance transparency and reproducibility, we provide the exact prompts used for dataset generation and visual reward feedback.

\begin{figure}[h]
    \centering
    \includegraphics[width=1.0\linewidth]{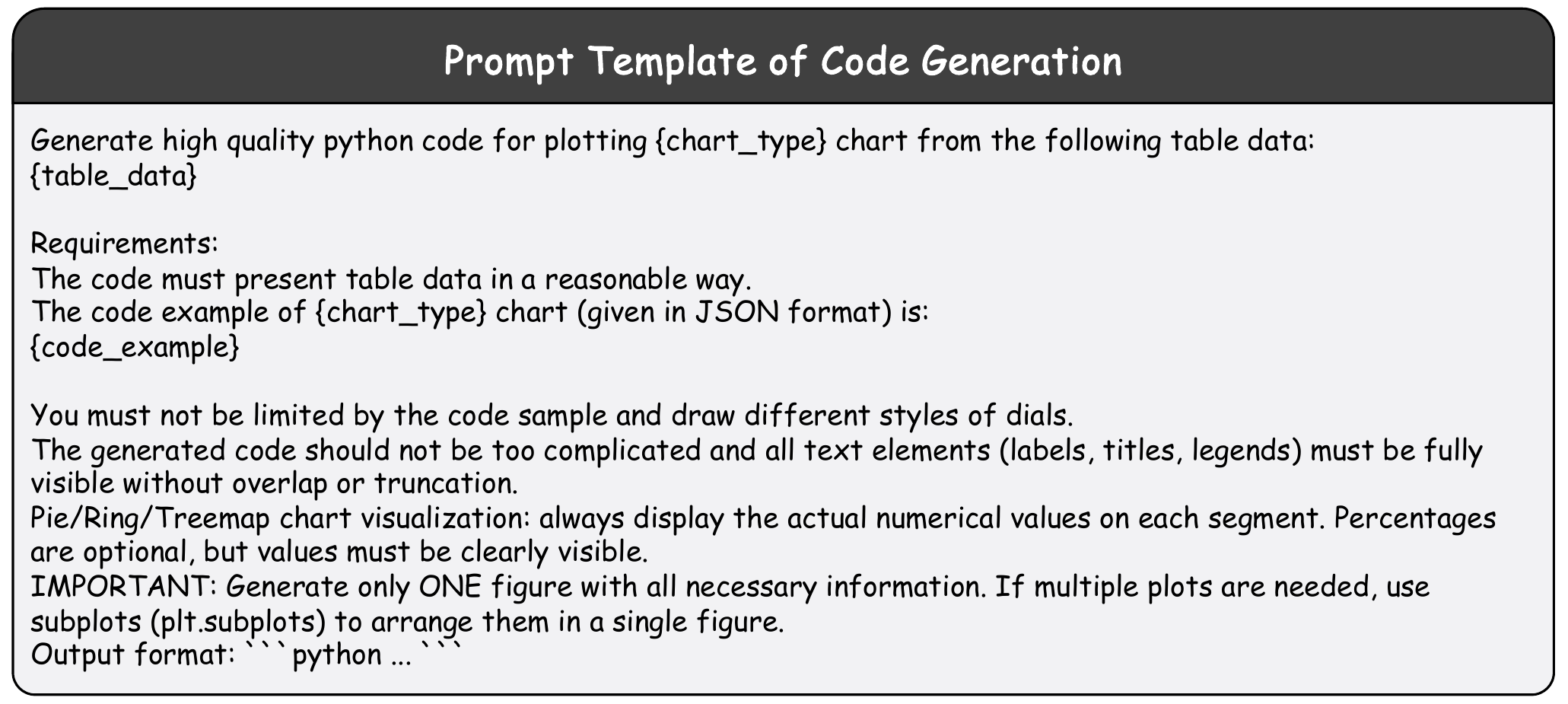}
    \vspace{-10pt}
    \caption{Prompt template for Chart2Code code generation.}
    \label{fig:prompts_code_gen}
    \vspace{-5pt}
\end{figure}

Figure~\ref{fig:prompts_code_gen} shows the prompt used for code generation. We use real table data as input, select one chart type from 24 predefined chart types, and sample code examples of the selected chart type to generate plotting code. Specifically, for chart types Pie, Ring, and Treemap, we require the model to display actual values on the image, as these chart types typically only visualize percentage values in their code, making it impossible to read the true values from the image.

\begin{figure}[h]
    \centering
    \includegraphics[width=1.0\linewidth]{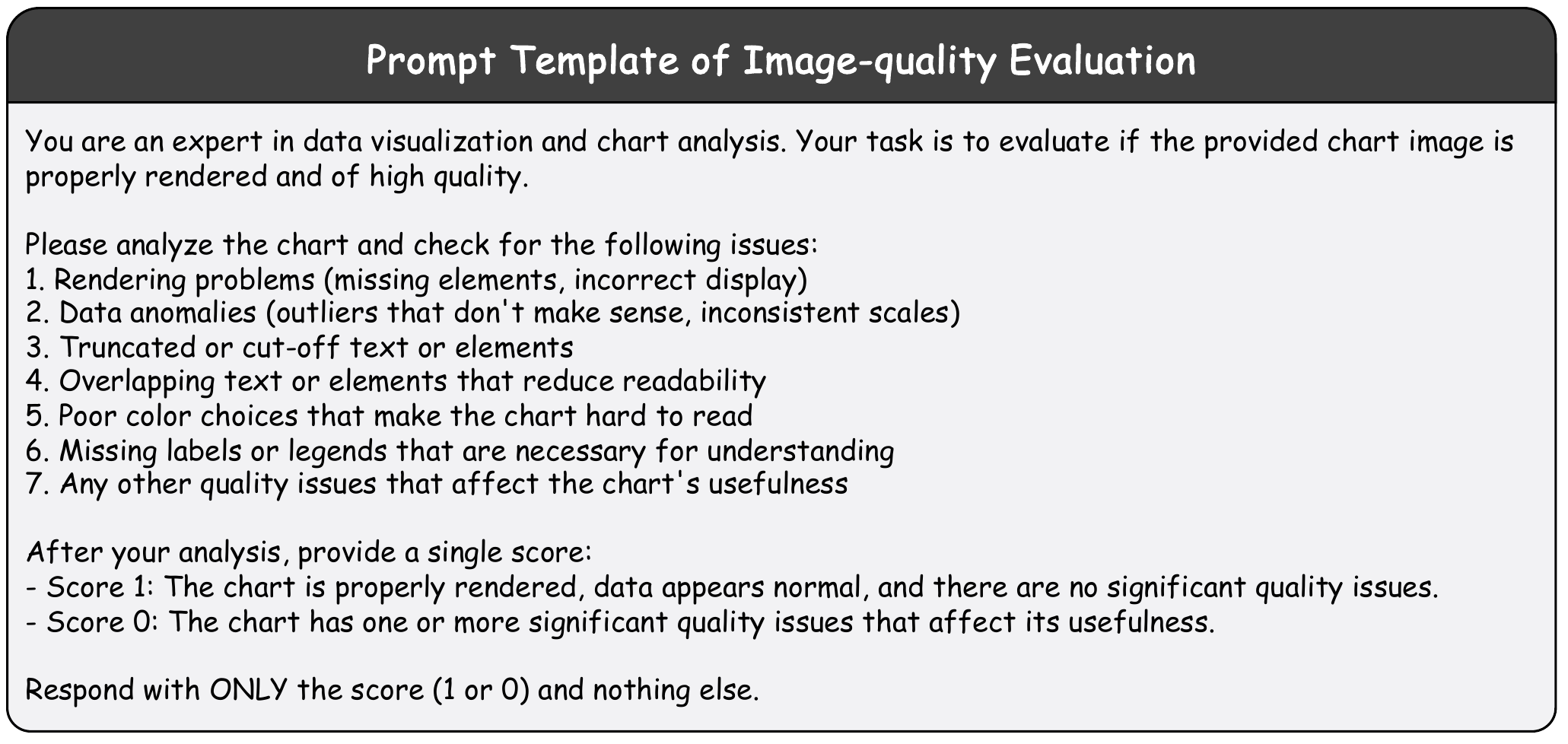}
    \vspace{-10pt}
    \caption{Prompt template for image-quality evaluation during RL data filtering.}
    \label{fig:prompts_img_eval}
    \vspace{-5pt}
\end{figure}

Figure~\ref{fig:prompts_img_eval} shows the prompt used for evaluating image quality. When processing RL data, we use this prompt to filter out visually low-quality samples, such as those with overlapping text or elements.

\begin{figure}[h]
    \centering
    \includegraphics[width=1.0\linewidth]{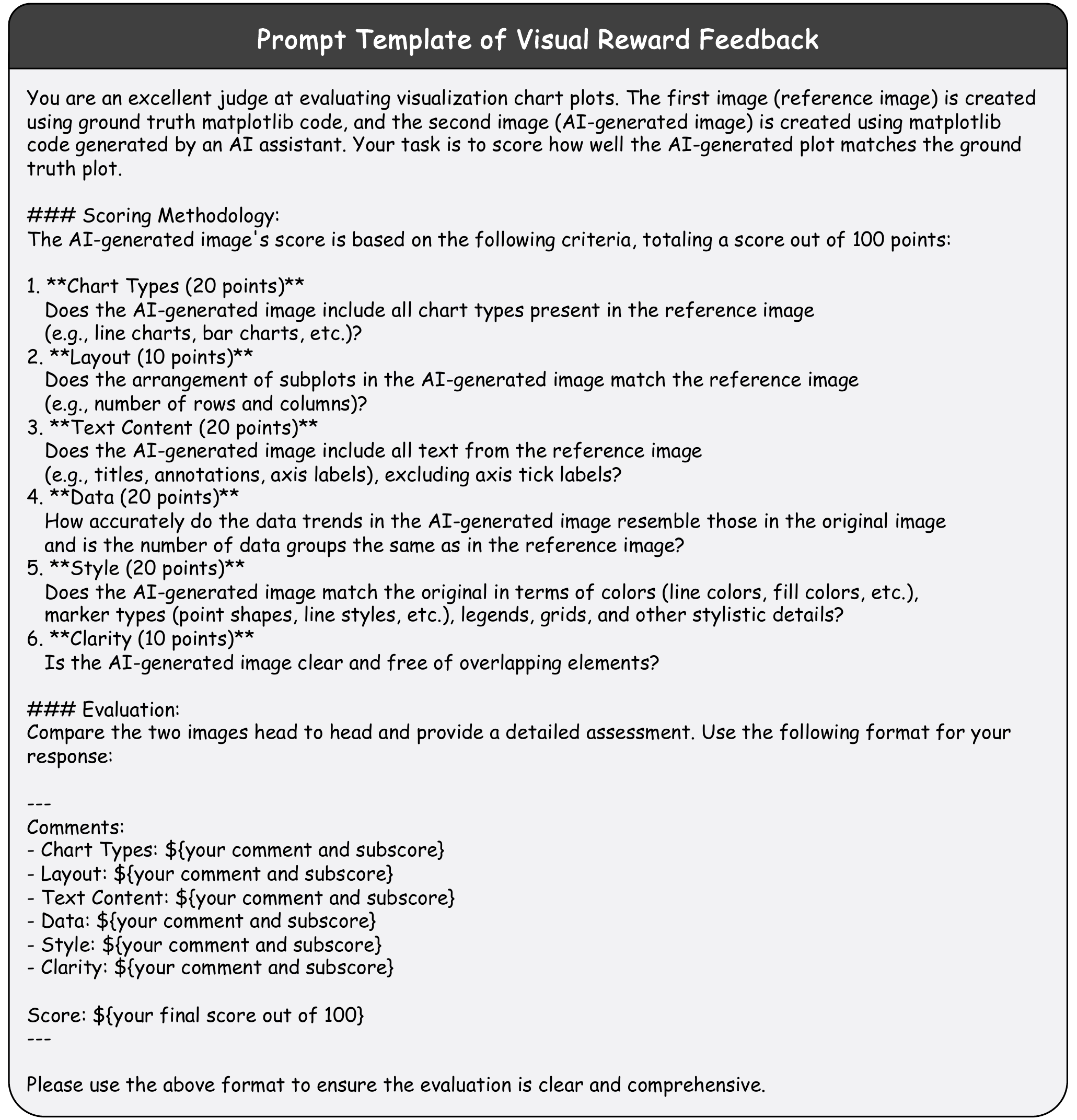}
    \vspace{-10pt}
    \caption{Prompt template for visual reward feedback during the RL training.}
    \label{fig:prompts_vis_reward}
    \vspace{-5pt}
\end{figure}

Figure~\ref{fig:prompts_vis_reward} shows the prompt used to calculate visual rewards. Following ChartMimic~\cite{yangchartmimic}, we assess the match between ground truth and model predictions across 6 dimensions, with this score normalized as the visual score and incorporated into the overall reward computation.


\section{The Use of Large Language Models (LLMs)}
In this work, we utilized LLMs primarily as writing assistance tools to polish language and improve clarity. They helped refine sentence structures and enhance readability throughout the manuscript. The core research ideas, experimental design, implementation, and analysis were all conducted by the authors, with LLMs serving only as editorial aids.



\end{document}